\documentclass{article} % For LaTeX2e

\usepackage{iclr2024_conference,times}

\usepackage{amsmath,amsfonts,bm}

\def\eqref#1{equation~\ref{#1}}
\def\1{\bm{1}}

\DeclareMathAlphabet{\mathsfit}{\encodingdefault}{\sfdefault}{m}{sl}
\SetMathAlphabet{\mathsfit}{bold}{\encodingdefault}{\sfdefault}{bx}{n}

\title{Language Modeling Is Compression}

\author{Gr{\'{e}}goire Del\'etang$^{*1}$ \And
Anian Ruoss$^{*1}$ \And
Paul-Ambroise Duquenne$^2$ \And
Elliot Catt$^1$ \And
Tim Genewein$^1$ \And
Christopher Mattern$^1$ \And
Jordi Grau-Moya$^1$ \And
Li Kevin Wenliang$^1$ \And
Matthew Aitchison$^1$ \And
Laurent Orseau$^1$ \And
Marcus Hutter$^1$ \And
Joel Veness$^1$}

\usepackage[utf8]{inputenc} % allow utf-8 input
\usepackage[T1]{fontenc}    % use 8-bit T1 fonts
\usepackage{hyperref}       % hyperlinks
\usepackage{url}            % simple URL typesetting
\usepackage{booktabs}       % professional-quality tables
\usepackage{amsfonts}       % blackboard math symbols
\usepackage{nicefrac}       % compact symbols for 1/2, etc.
\usepackage{microtype}      % microtypography
\usepackage{xcolor}         % colors

\usepackage{algorithm,algpseudocode}
\usepackage{amsmath}
\usepackage{amssymb}
\usepackage{bm}
\usepackage{bbm}
\usepackage[scaled=0.9]{beramono}
\usepackage{chngcntr}
\usepackage{dsfont}
\usepackage{glossaries}
\usepackage{graphicx}
\usepackage{listings} 
\usepackage{multirow}
\usepackage{multicol}
\usepackage[super]{nth}
\usepackage{siunitx}
\usepackage{subcaption}
\usepackage{wrapfig}
\usepackage{tikz}
\usepackage{pgfplots}
\pgfplotsset{compat=newest}
\usepackage{graphicx}
\usepackage{multicol}

\usepackage[outdir=./figures/]{epstopdf}

\usepackage[capitalize]{cleveref}
\newcommand{\eg}{e.g., }

\newcommand{\ie}{i.e., }

\newcommand{\versus}{vs.\ }

\iclrfinalcopy

\begin{document}

\def\thefootnote{*}\footnotetext{Equal contribution. $^1$Google DeepMind. $^2$Meta AI \& Inria. Correspondence to \{gdelt, anianr\}@google.com.}\def\thefootnote{\arabic{footnote}}

\maketitle

\begin{abstract}
    It has long been established that predictive models can be transformed into lossless compressors and vice versa.
    Incidentally, in recent years, the machine learning community has focused on training increasingly large and powerful self-supervised (language) models.
    Since these large language models exhibit impressive predictive capabilities, they are well-positioned to be strong compressors.
    In this work, we advocate for viewing the prediction problem through the lens of compression and evaluate the compression capabilities of large (foundation) models.
    We show that large language models are powerful general-purpose predictors and that the compression viewpoint provides novel insights into scaling laws, tokenization, and in-context learning. For example, Chinchilla 70B, while trained primarily on text, compresses ImageNet patches to 43.4\% and LibriSpeech samples to 16.4\% of their raw size, beating domain-specific compressors like PNG (58.5\%) or FLAC (30.3\%), respectively. Finally, we show that the prediction-compression equivalence allows us to use any compressor (like gzip) to build a conditional generative model.
\end{abstract}

\section{Introduction}\label{sec:introduction}

Information theory and machine learning are inextricably linked and have even been referred to as ``two sides of the same coin''~\citep{mackay2003information}.
One particularly elegant connection is the essential equivalence between probabilistic models of data and lossless compression.
The source coding theorem~\citep{shannon1948mathematical} is the fundamental theorem describing this idea, \ie the expected message length in bits of an optimal entropy encoder is equal to the negative $\log_2$-likelihood of the statistical model.
In other words, maximizing the $\log_2$-likelihood (of the data) is equivalent to minimizing the number of bits required per message.
Indeed, lossless compression with a probabilistic model can be achieved in a variety of different ways, including Huffman coding~\citep{huffman1952method}, arithmetic coding~\citep{pasco1977source, rissanen1976generalized}, and asymmetric numeral systems~\citep{duda2009asymmetric}.

\begin{figure}
    \begin{center}
        \includegraphics[trim={9.5mm 0 0 0},clip,width=\textwidth]{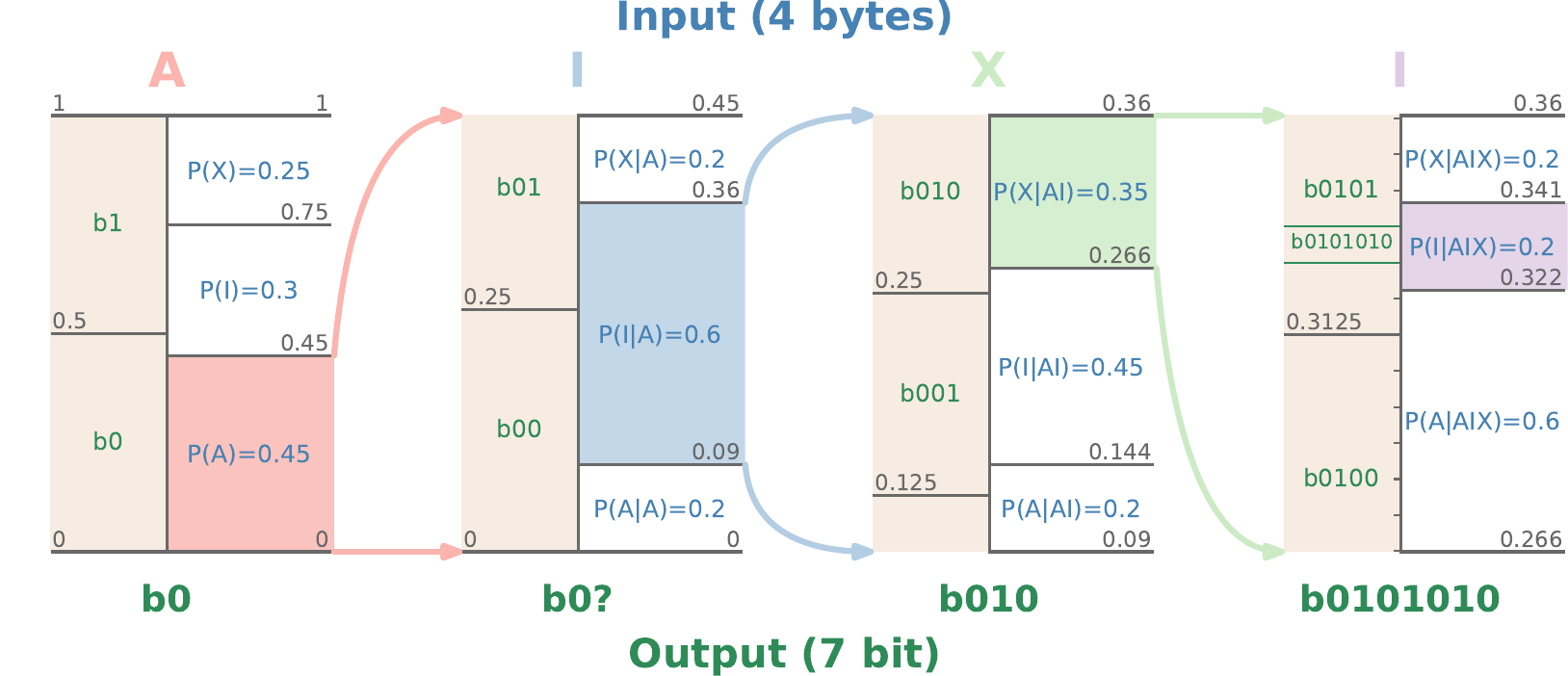}
    \end{center}
    \caption{
        Arithmetic encoding of `AIXI' with a probabilistic model $P$ (blue) resulting in the binary code `b0101010' (green). 
        We iteratively divide the real interval $I = [0,1)$ according to the model's (conditional) probabilities and select the sub-interval corresponding to the observed symbol (\eg $I = [0,0.45)$ for $P(A)$).
        We further refine $I$ for each input symbol (indicated by the arrows), \eg $I = [0.09, 0.36)$ for $P(I|A)$.
        To determine the encoded output, we iteratively split $[0,1)$ in half and assign a binary code to each sub-interval (shaded red areas).
        At every step we can output the binary code if $I$ is fully contained in the corresponding binary interval (\eg `b0' for `A', but not for `AI' as it could be `b00' or `b01').
        At the end of the input, the code is `b0101', which cannot be uniquely decoded ($P(A|AIX)$, $P(I|AIX)$, $P(X|AIX)$ all overlap with `b0101').
        Thus, we further refine the binary code until its binary interval is fully contained in $I$ (all calculations in \cref{sec:arithmetic-coding}).
    }
    \label{fig:overview}
\end{figure}

Arithmetic coding, in particular, is known to be optimal in terms of coding length, meaning that the overall compression performance depends on the capabilities of the probabilistic model (see \cref{fig:overview} for an overview of arithmetic coding).
Incidentally, in recent years, large pre-trained Transformers~\citep{vaswani2017attention}, so-called \emph{foundation models}~\citep{bommasani2021opportunities}, have proven to be highly successful across a wide range of predictive tasks \citep{bubeck2023sparks, rae2021scaling} and are thus promising candidates for use with arithmetic coding.
Indeed, Transformer-based compression with arithmetic coding has produced state-of-the-art results both in the online~\citep{bellard2021nncp, mao2022trace} and offline settings~\citep{valmeekam2023llmzip}.
In the online setting, a pseudo-randomly initialized model is directly trained on the stream of data that is to be compressed, while the offline setting, which we consider in our work, trains the model on an external dataset before employing it to compress a (potentially different) data stream.
Consequently, offline compression is performed \emph{in-context}, with a fixed set of model parameters.
Transformers have demonstrated impressive in-context learning abilities~\citep{laskin2023incontext, brown2020language, wei2022chain, genewein2023memory} and are thus ideally suited for offline compression.

The context length is a key limiting factor in offline compression, as it dictates the maximum number of bytes a model can compress at a time.
Transformers can only compress a few kilobytes (each ``token'' being coded with 2 or 3 bytes), while requiring a lot of compute.
Correspondingly, many challenging predictive tasks (\eg algorithmic reasoning or long-term memory) require long contexts~\citep{deletang2023neural}, and thus extending these models' context lengths is a key challenge which is gaining increased attention~\citep{zaheer2020big, guo2022longt5, bulatov2023scaling}.
The in-context compression view provides insights into the failure modes of current foundation models.

\paragraph{This Work}

We advocate for using (lossless) compression to study foundation models.
To that end, we conduct an extensive empirical investigation of the offline (in-context) compression capabilities of large language models, with the rationale that they have become readily available~\citep{touvron2023llama, touvron2023llama2} and can thus be used for compression without the training overhead.
We empirically demonstrate that these models, while (meta-)trained primarily on text, achieve competitive compression rates across different data modalities, outperforming domain-specific standard compressors (not accounting for model parameter size). 
Moreover, we shed new light on scaling laws~\citep{kaplan2020scaling}, showing that they also hold true for compression but that measuring the adjusted compression rates instead of the log loss adds a twist: Scaling beyond a certain point will deteriorate the compression performance since the model parameters need to be accounted for in the compressed output.
Finally, we advocate for framing (self-supervised) prediction through the lens of compression as it encompasses generalization: a model that compresses well generalizes well~\citep{hutter2006prize}.

\paragraph{Contributions}

We empirically study the lossless compression capabilities of foundation models:

\begin{itemize}
    \item 
    We review how to compress with predictive models via arithmetic coding and call attention to the connection between current language modeling research and compression.
    \item We show that large language models achieve impressive compression rates (disregarding model parameter size) on modalities other than text.
    For example, Chinchilla 70B achieves compression rates of 43.4\% on ImageNet patches and 16.4\% on LibriSpeech samples, beating domain-specific compressors like PNG (58.5\%) or FLAC (30.3\%), respectively.
    \item We revisit scaling laws, showing that the dataset size provides a hard limit on model size in terms of compression performance and that model scaling is not a silver bullet.
    \item We leverage the compression-prediction equivalence to employ compressors as generative models and visually illustrate the performance of the underlying compressor.
    \item We demonstrate that tokenization, which can be viewed as a pre-compression, does, in general, not improve compression performance, but allows models to increase the information content in their context and is thus generally employed to improve prediction performance.
\end{itemize}

\section{Background}\label{sec:background}

In this section, we review the necessary background on information theory and its relation to likelihood maximization.
To that end, we consider streams of data $x_{1:n} := x_1 x_2 \ldots x_n \in \mathcal{X}^n$ of length $n$ from a finite set of symbols $\mathcal{X}$.
We write $x_{\leq j}=x_{<j+1}:=x_{1:j}$ for $j\leq n$ and denote the empty string as $\epsilon$.
Finally, we denote the concatenation of two strings $s$ and $r$ by $sr$.

\paragraph{Coding Distributions} 

A coding distribution $\rho$ is a sequence of probability mass functions $\rho_n : \mathcal{X}^n \mapsto (0, 1]$, which for all $n \in \mathbb{N}$ satisfy the constraint that $\rho_n(x_{1:n}) = \sum_{y \in \mathcal{X}}\rho_{n+1}(x_{1:n}y)$ for all $x_{1:n} \in \mathcal{X}^n$, with the base case $\rho_0(\epsilon) := 1$.
From here on out, whenever the meaning is clear from the argument to $\rho$, we drop the subscript on $\rho$.
Under this definition, the conditional probability of a symbol $x_n$ given previous data $x_{< n}$ is defined as $\rho(x_n \mid x_{< n}) := \rho(x_{1:n}) / \rho(x_{< n})$, with the familiar chain rules $\rho(x_{1:n}) = \prod_{i=1}^n \rho(x_i \mid x_{< i})$ and $\rho(x_{j:k} \mid x_{< j}) = \prod_{i = j}^k \rho(x_i \mid x_{<i})$ following.

\paragraph{Lossless Compression}

The goal of lossless compression is to encode a stream of symbols $x_{1:n}$ sampled from a coding distribution $\rho$ into a bitstream of minimal (expected) length, while ensuring that the original data sequence is recoverable from the bitstream.
To that end, we use a binary source code $c : \mathcal{X}^* \mapsto \{0, 1\}^*$, which assigns to each possible data sequence $x_{1:n}$ a binary code word $c(x_{1:n})$ of length $\ell_c(x_{1:n})$ (in bits).
Thus, the aim is to minimize the expected bits per sequence $L := \mathrm \mathbb{E}_{x \sim \rho}[\ell_c(x)]$, \ie encoding rare sequences with more bits and frequent sequences with fewer bits.
Shannon's source coding theorem establishes the limit on possible data compression as $L \geq H(\rho)$ for any possible code, where $H(\rho) := \mathbb{E}_{x \sim \rho} [- \log_2 \rho(x)]$ is the Shannon entropy~\citep{shannon1948mathematical}.

\paragraph{Arithmetic Coding}

Given a coding distribution $\rho$ and a sequence $x_{1:n}$, arithmetic coding~\citep{pasco1977source, rissanen1976generalized} constructs a code with almost optimal length.
It directly connects coding and compression with prediction and modeling: compressing well means modeling well in a log-loss sense and vice-versa.
Assuming infinite precision for the arithmetic operations involved, the arithmetic code has length $-\lceil\log\rho(x_{1:n})\rceil + 1$ bits, whereas the optimal code length is $-\log\rho(x_{1:n})$ bits.
A practical implementation that is subject to $B$ bit precision adds further $O(n 2^{-B})$ bits~\citep{howard1991analysis}, which is negligible for 32- or 64-bit arithmetic.
In the following we consider infinite precision arithmetic coders and refer to \citet{witten1987arithmetic} for the finite-precision implementation.

\paragraph{Arithmetic Encoder}

The arithmetic code of a sequence $x_{1:n}$ is the binary representation of a number $\lambda\in[0, 1)$.
We identify $\lambda$ by narrowing down an interval that encloses $\lambda$ step by step (maintaining a growing prefix of the binary representation of $\lambda$ throughout the process).
Initially, this interval is $I_0 = [0, 1)$.
In step $k > 0$ (\ie encoding $x_k$), we first partition the previous interval $I_{k-1} = [l_{k-1}, u_{k-1})$ into $N$ sub-intervals $\tilde I_k(x_1), \tilde I_k(x_2), \dots$, one for each letter from $\mathcal X = \{x_1, x_2, \dots, x_N\}$.
The size of sub-interval $\tilde I_k(y)$ that represents letter $y$ is $(u_{k-1} - l_{k-1})\cdot \rho(y\mid x_{<k})$.
Formally, we define
\begin{equation}
    \tilde I_k(x) := \left[l_{k-1} + (u_{k-1} - l_{k-1})\cdot\sum_{y<x} \rho(y\mid x_{<k}),\quad l_{k-1} + (u_{k-1} - l_{k-1})\cdot\sum_{y\leq x} \rho(y\mid x_{<k}) \right)
    \text,
\end{equation}
assuming a strict order on $\mathcal X$.
To encode $x_k$ we proceed with its corresponding interval, \ie $I_k = \tilde I_k(x_k)$.
Finally, we choose $\lambda\in I_n$ with the shortest binary representation in the terminating interval $I_n$ and use that binary representation to encode $x_{1:n}$.
\cref{fig:overview} illustrates this process.

\paragraph{Arithmetic Decoder}

Given $\lambda$ and $\rho$ decoding the $k$-th letter is easy: Starting with $I_0 = [0, 1)$, find $y$ such that $\lambda\in\tilde I_k(y)$ to decode $x_k=y$, then set $I_k=\tilde I_k(x_k)$ and proceed with the $k\!+\!1$-st letter.

\paragraph{Likelihood Maximization}

In practice, the source distribution $\rho$ is usually unknown and is instead estimated with a parametric probabilistic model $\hat{\rho}$.
Thus, instead of achieving code length $-\sum_{i = 1}^n \log_2\rho(x_i \mid x_{<i})$ for the sequence $x_{1:n}$, we obtain the suboptimal length $-\sum_{i = 1}^n \log_2\hat{\rho}(x_i \mid x_{<i})$.
As a result, the expected (suboptimal) number of bits is the \emph{cross-entropy}:
\begin{equation}
    H(\rho, \hat{\rho}) := \mathbb{E}_{x \sim \rho}\left[\sum_{i=1}^n-\log_2 \hat{\rho}(x_i \mid x_{<i})\right].
    \label{eq:cross-entropy}
\end{equation}
Thus, we can minimize the expected length of the encoded data stream with symbols distributed according to $\rho$ by minimizing the cross-entropy with respect to some $\hat{\rho}$, which is equivalent to likelihood maximization~\citep{mackay2003information}.
However, \cref{eq:cross-entropy} is exactly the same objective used to train current foundation models, \ie the $\log$-loss. 
Thus, minimizing the $\log$-loss is equivalent to minimizing the compression rate of that model used as a lossless compressor with arithmetic coding, \ie current language model training protocols use a maximum-compression objective.

\paragraph{Compression-Based Sequence Prediction}

Analogous to how a predictive distribution can be used for lossless compression via arithmetic coding (described above), any compressor can be employed for sequence prediction~\citep{frank2000text}.
The main idea is to define $\rho(x_{1:n})$ as the coding distribution $2^{-\ell_c(\cdot)}$, where $\ell_c(x_{1:n})$ is the length of sequence $x_{1:n}$ when encoded with compressor $c$ (\eg gzip).
We thus recover the conditional distribution $\rho(x_i \mid x_{< i})$ by computing $2^{\ell_c(x_{<i}) - \ell_c(x_{<i}x_i) }$, for all $x_i$.

\paragraph{Universal Coding}

Above we discussed optimal (arithmetic) coding with respect to data sampled from a fixed distribution $\rho$.
In contrast, universal (optimal) source coding with respect to all computable sampling distributions can, in theory, be achieved by choosing $\ell_c(x_{1:n})$ as the Kolmogorov complexity of $x_{1:n}$~\citep{kolmogorov1998table, li2019introduction}.
For this choice, the conditional distribution described above is universally optimal over $x_{<i}$, recovering the Solomonoff predictor~\citep{solomonoff1964formal1, solomonoff1964formal2, rathmanner2011philosophical}.
The Solomonoff predictor is a Bayesian mixture of \emph{all} predictors that can be programmed in a chosen Turing-complete programming language.
More precisely, for a predictor $q$ of program-length $\ell_c(q)$ bits,
the Solomonoff predictor assigns a prior weight of $2^{-\ell_c(q)}$ to predictor $q$.
That is, if $\mathcal{Q}$ is the set of all predictors that can be programmed and computed, the Solomonoff predictor assigns probability $S(x_{1:n})=\sum_{q\in{\mathcal{Q}}} 2^{-\ell_c(q)} q(x_{1:n})$ to a sequence $x_{1:n}$.
Therefore, $S(x_{1:n})\geq 2^{-\ell_c(q)} q(x_{1:n})$ for all $q\in\mathcal{Q}$, and thus $-\log_2 S(x_{1:n}) \leq -\log_2 q(x_{1:n}) + \ell_c(q)$.
Observe that $\ell_c(q)$ is a constant of $q$ that is independent of the sequence length.
Therefore, compressing optimally is equivalent to predicting optimally and vice versa~\citep{hutter2005universal}.

\section{Experimental Evaluation}\label{sec:experiments}

Here, we evaluate foundation models' (in-context) compression capabilities (details in \cref{sec:experimental-details} and
code at {\small \url{https://github.com/google-deepmind/language_modeling_is_compression}}).

\paragraph{Compressors}

We compare our arithmetic coding-based language model compressors to two competitive general-purpose lossless compressors: gzip~\citep{deutsch1996gzip} and its improvement LZMA2~\citep{pavlov20197z}, used by the 7zip software.
Both are based on Huffman coding~\citep{huffman1952method} and the Lempel-Ziv-Welch algorithm~\citep{welch1984technique}.
We also consider specialized lossless compressors for image and audio data, \ie  PNG~\citep{boutell1997png} and FLAC~\citep{coalson2008flac}, respectively.
Finally, we evaluate two types of language models (of different sizes) with arithmetic coding: vanilla decoder-only Transformers~\citep{vaswani2017attention}, which we train on the enwik8 dataset, and the pretrained Llama 2~\citep{touvron2023llama2} and Chinchilla~\citep{hoffmann2022training}.

\subsection{Datasets}
\label{sec:experiments:datasets}

We consider datasets of three different modalities, text, image, and audio, which have (a priori) very different biases for compression and thus provide a good testbed for evaluating a compressor's general capabilities.
To render the results comparable across modalities, all our datasets are 1GB.

A key question is how to reconcile the different context lengths $C$ of the compressors we consider.
Transformers are restricted to short contexts (2048 ``tokens'', coded over 1 byte for our trained transformers, and 4 bytes for the pretrained models), while gzip uses a maximum context of 32 kilobytes, and LZMA2 has a virtually ``infinite'' context length.
Having a longer context allows a compressor to exploit more sequential dependencies to achieve a better compression rate.
For compressors with finite contexts, there are two approaches to compress sequences that are longer than the context length: (i) slide the compressor byte by byte, thus always processing a history of the previous $C - 1$ bytes when compressing a new byte, and (ii) chunk the data stream into $S$ sequences of $C$ bytes and evaluate the in-context compression (without any history) averaged across batches.
For Transformers, we consider the latter approach since sliding would increase their (already very long) running time by a factor of $S$.
Therefore, we chunk all datasets into sequences of $2048$ bytes and feed them to the compressors one-by-one.
However, since classical compressors usually include a header in their compressed output, which can be larger than the compressed data in some cases, we only count it once for all batches.
Moreover, since chunking deteriorates the performance of classical compressors, which have context lengths $C \gg 2048$, we also report their compression rates on the unchunked datasets.
We consider the following datasets:

\paragraph{enwik9}

The enwik9 dataset~\citep{hutter2006prize} consists of the first \num{1000000000} (1 billion) bytes of the English Wikipedia XML dump on March 3rd, 2006 and is typically used to measure a model's ability to compress data.
It is an extension of the enwik8 dataset that only contains the first 100 million bytes.
We train our vanilla Transformer models on enwik8, but evaluate on both enwik8 and enwik9 (to evaluate the out-of-distribution compression performance).
While enwik8 is included in enwik9, it only represents the first 10\% and thus still constitutes a significant distribution shift.

\paragraph{ImageNet}

The ImageNet dataset~\citep{russakovsky2015imagenet} contains \num{14197122} annotated images from the WordNet hierarchy.
Since 2010, the dataset has been used in the ImageNet Large Scale Visual Recognition Challenge (ILSVRC), a benchmark in image classification and object detection.
We extract contiguous patches of size $32 \times 64$ from all images, flatten them, convert them to grayscale (so that each byte represents exactly one pixel) to obtain samples of 2048 bytes.
We then concatenate \num{488821} of these patches, following the original dataset order, to create a dataset of 1 GB.

\paragraph{LibriSpeech}

LibriSpeech~\citep{panayotov2015librispeech} contains roughly \num{1000} hours of 16kHz English speech data derived from audiobooks of the LibriVox project that has been segmented and aligned.
We chunk the samples into 2048 bytes and gather \num{488821} such chunks into a dataset of size 1 GB.

\begin{table}
  \caption{
    Compression rates (compressed size / raw size) on different datatsets (lower is better).
    The raw compression rate does not take the parameter size into account for the neural models, while the adjusted compression rate considers the parameter size part of the compressed size.
    All datasets are of size 1GB.
    Random data is used as a baseline and should not be compressible.
    Transformer, Llama 2, and Chinchilla are predictive models, which we use with arithmetic coding to obtain lossless compressors.
    We train the Transformer models from scratch on enwik8, while the Chinchilla models are pretrained on large text datasets.
    Transformers trained on enwik overfit to that data modality, while large language models are good compressors for various data types.
  }
  \label{tab:compression-rates}
  \begin{center}
    \resizebox{\textwidth}{!}{\begin{tabular}{@{}llcrrrrcrrrr@{}}
    \toprule
    & && \multicolumn{4}{c}{Raw Compression Rate (\%)} && \multicolumn{4}{c}{Adjusted Compression Rate (\%)} \\
    \cmidrule{4-7}
    \cmidrule{9-12}
    \textbf{Chunk} & \textbf{Compressor} && \textbf{enwik9} & \textbf{ImageNet} & \textbf{LibriSpeech} & \textbf{Random} && \textbf{enwik9} & \textbf{ImageNet} & \textbf{LibriSpeech} & \textbf{Random} \\
    \midrule
    \multirow{4}{*}{$\infty$} & gzip             &&         32.3 &          70.7 &          36.4 & 100.0 &&          32.3 &          70.7 &          36.4 &   100.0 \\
                              & LZMA2            &&         23.0 &          57.9 &          29.9 & 100.0 &&          23.0 & \textbf{57.9} &          29.9 &   100.0 \\
                              & PNG              &&         42.9 &          58.5 &          32.2 & 100.0 &&          42.9 &          58.5 &          32.2 &   100.0 \\
                              & FLAC             &&         89.5 &          61.9 &          30.9 & 107.8 &&          89.5 &          61.9 &          30.9 &   107.8 \\
    \midrule                                                                
    \multirow{11}{*}{$2048$}  & gzip             &&         48.1 &          68.6 &          38.5 & 100.1 &&          48.1 &          68.6 &          38.5 &   100.1 \\
                              & LZMA2            &&         50.0 &          62.4 &          38.2 & 100.0 &&          50.0 &          62.4 &          38.2 &   100.0 \\
                              & PNG              &&         80.6 &          61.7 &          37.6 & 103.2 &&          80.6 &          61.7 &          37.6 &   103.2 \\
                              & FLAC             &&         88.9 &          60.9 &          30.3 & 107.2 &&          88.9 &          60.9 & \textbf{30.3} &   107.2 \\
    \cmidrule{2-12}                                              
                              & Transformer 200K &&         30.9 &         194.0 &         146.6 & 195.5 &&          30.9 &         194.0 &         146.6 &   195.5 \\
                              & Transformer 800K &&         21.7 &         185.1 &         131.1 & 200.1 &&          21.9 &         185.3 &         131.3 &   200.3 \\
                              & Transformer 3.2M &&         17.0 &         215.8 &         228.2 & 224.0 && \textbf{17.7} &         216.5 &         228.9 &   224.7 \\
    \cmidrule{2-12}                           
                              & Llama 2 (7B)     &&          8.9 &          53.4 &          23.1 & 103.2 &&        1408.9 &        1453.4 &        1423.1 &  1503.2 \\
    \cmidrule{2-12}                           
                              & Chinchilla 1B    &&         11.3 &          62.2 &          24.9 & 108.8 &&         211.3 &         262.2 &         224.9 &   308.8 \\
                              & Chinchilla 7B    &&         10.2 &          54.7 &          23.6 & 101.6 &&        1410.2 &        1454.7 &        1423.6 &  1501.6 \\
                              & Chinchilla 70B   && \textbf{8.3} & \textbf{48.0} & \textbf{21.0} & 100.8 &&       14008.3 &       14048.0 &       14021.0 & 14100.8 \\
    \bottomrule
\end{tabular}
}
  \end{center}
\end{table}

\subsection{Comparing Compression Rates}
\label{sec:experiments:compression-rates}

\cref{tab:compression-rates} shows the compression rates for all compressors and datasets.
We show both the raw compression rate, which does not take the model size (in bytes) into account, as well as the adjusted rate, which does.
The size of the Python program for classical compressors is very small (a few kilobytes at most) and thus barely affects the compression rate.
In contrast, language models suffer a huge loss in compression rate due to their large size, which cannot be offset when compressing only 1GB of data.
We encode each neural network parameter with 2 bytes, using a float16 representation since quantizing weights to this level does not significantly affect performance~\citep{tao2022compression} and is standard for model inference.
Note that further compressing the float16 parameters using classical compressors does not significantly reduce their size (we obtained rates of 92.2\% and 89.1\% on a 38M parameter Transformer with gzip and LZMA2, respectively).
We only consider the offline setting, which computes the adjusted compression rate using a two-part code (i.e., it adds the model size to the $\log$-loss of the data).
In contrast, prequential (online) coding would provide an alternative view on adjusted compression by computing the adjusted compression rate as the $\log$-loss plus the size of the training script (not the model parameters). 
Prequential coding leads to better compression with overparametrized neural networks~\citep{blier2018description}, but it requires training the model online both during encoding and decoding (which is very costly for our models). 

\paragraph{Foundation Models Are General-Purpose Compressors}

A lossless compressor induces an injective function over bit sequences, meaning that we cannot compress all sequences equally well (by the pigeonhole principle).
Consequently, in practice, compressors are often tailored to a particular setting, \eg FLAC for audio or PNG for images, and thus fail to compress other data modalities well (see \cref{tab:compression-rates}).
In contrast, general-purpose compressors, such as gzip, offer good performance on a wide range of data sources.
Surprisingly, large language models, while trained primarily on text, also appear to be general-purpose compressors, as they outperform all other compressors, even on image and audio data (see \cref{tab:compression-rates}).
Note that these models have not been trained on this kind of data: for Chinchilla, Appendix A.\ of \citet{hoffmann2022training} states that the training dataset consists of a mix of internet text data (Wikipedia, websites, github) and books.
However, it is still possible (but unlikely) that some images or audio samples were encoded into text on some websites.
Thus, pretrained models achieve their impressive compression performance by conditioning a (meta-)trained model to a particular task at hand via in-context learning~\citep{genewein2023memory}.
In contrast, smaller Transformers, trained manually on enwik8, only achieve good compression rates on similar Wikipedia data, \ie enwik9.
However, larger models' stronger in-context compression (or in-context learning) comes at a price: the number of parameters, which has to be offset with increasingly large data sources when computing the adjusted compression rate (see \cref{sec:experiments:tradeoff}).

\begin{figure}
    \begin{minipage}{0.55\linewidth}
        \begin{center}
            \includegraphics[width=\textwidth]{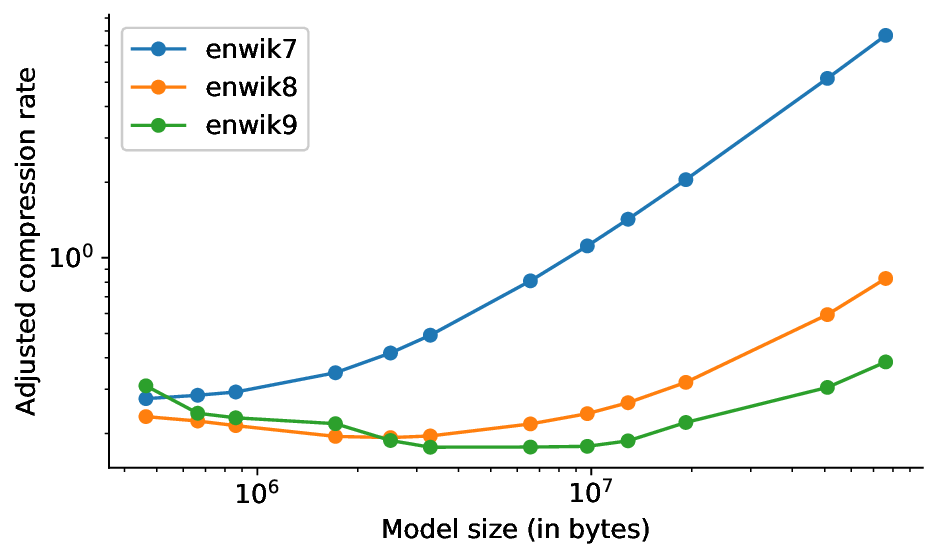}
        \end{center}
        \captionof{figure}{
            Adjusted compression rates (compressed size / raw size) for Transformers of different sizes, trained on enwik8 and evaluated on enwik (both axes are logarithmic).
            Here, the compressed size does not only consider the size of the compressed output (roughly equal to the $\log$-loss) but also the model size, which causes all curves to increase at some point.
            Every dataset gives rise to an optimal model size, with a good trade-off between performance (the size of the compressed data) and cost of the model (the number of parameters).
            The larger the dataset, the more parameters we can afford.
        }
        \label{fig:tradeoff}
    \end{minipage}
    \hfill
    \begin{minipage}{0.4\linewidth}
        \captionof{table}{
            Raw compression rates (compressed size / raw size) on enwik9 for Transformers trained on enwik8 with different tokenizers, ASCII and byte-pair encoding (BPE), with various vocabulary sizes.
            Transformers compress better with simpler tokenizers.
            However, larger vocabulary sizes reduce the length of the sequence more, meaning more information can be packed into the context.
         }
         \label{tab:tokenizers}
        \begin{center}
            \resizebox{\textwidth}{!}{\begin{tabular}{@{}lcrrr@{}}
    \toprule
    && \multicolumn{3}{c}{Raw Compression Rate (\%)} \\
    \cmidrule{3-5}
    \textbf{Tokenizer} && \textbf{200K} & \textbf{6.4M} & \textbf{38M} \\
    \midrule
    ASCII   && 22.9          & \textbf{13.6} & \textbf{6.4} \\
    BPE 1K  && 25.4          & 14.8          & 6.9 \\
    BPE 2K  && 25.6          & 15.7          & 7.4 \\
    BPE 5K  && 23.1          & 17.1          & 8.7 \\
    BPE 10K && 21.3          & 17.0          & 8.9 \\
    BPE 20K && \textbf{19.3} & 16.4          & 9.0 \\
    \bottomrule
\end{tabular}}
        \end{center}
    \end{minipage}
\end{figure}

\subsection{Optimal Model-Dataset Size Tradeoff}
\label{sec:experiments:tradeoff}

As shown in \cref{tab:compression-rates}, foundation models incur a huge cost in compression rates when accounting for their size, which is in the order of hundreds of GBs for billions of parameters.
In theory, if the dataset is infinite, we can ignore the model's size since it is insignificant compared to the size of the dataset.
However, in practice, a foundation model can only achieve non-trivial (adjusted) compression rates when evaluated on datasets in the order of TBs (or more).
Since this is infeasible under reasonable hardware constraints, we instead investigate the optimal model size with smaller Transformers that we train on enwik8.
Recall that the model size (in bytes) is twice the number of (float16) parameters.

\cref{fig:tradeoff} visualizes the adjusted compression rate for vanilla Transformers of different sizes for enwik.
We observe that larger models achieve better compression rates on larger datasets, justifying recent trends in model scaling~\citep{kaplan2020scaling}.
However, they achieve worse rates on smaller datasets, indicating that scaling laws are, in fact, dependent on the size of the test set.
That is, for each dataset, the model sizes reach a critical point, after which the adjusted compression rate starts to increase again as the number of parameters overweighs the size of the dataset.
Note that we evaluate offline compression, \ie we do not necessarily compress the data the model was trained on, meaning that the results on enwik7 and enwik8 are in-distribution, while enwik9 is (partially) out-of-distribution.

\subsection{Compressors as Generative Models}
\label{sec:experiments:generation}

In \cref{sec:experiments:compression-rates} we showed that any predictor can be employed as a compressor.
Here, following \cref{sec:background}, we empirically demonstrate the opposite direction, \ie that compressors can be used as a sequence prediction model, establishing our main claim that ``language modeling is compression''.
We compute the length~$\ell_c$ of the compressed sequence~$c(x_{<i}b)$ for all possible $b \in \mathcal{X}$ to get the probabilities {$\hat{\rho}(b \mid x_{<i}) = 2^{\ell_c(x_{<i}) - \ell_c(x_{<i}b)}$}.
This can straightforwardly be extended to sampling a whole continuation autoregressively by appending the last output to the sequence and iterating.

Theoretically, there is no strong guarantee that a good compression rate leads to ``good'' autoregressive samples.
However, empirically it has been shown that better sequence prediction (\ie lower $\log$-loss) often leads to better generation~\citep{rae2021scaling, brown2020language}.
Nevertheless, in autoregressive sampling small errors often accumulate, which can lead to samples that diverge from the ground-truth distribution.
Also, this standard sampling technique only looks one step into the future, and can be biased: gzip, for instance, builds an internal dictionary of 'tokens', which will be compressed using their indexes. Extending the sequence $x_{<i}$ with one of these tokens will lead to a good compression rate, but will be omitted as it can be longer than one byte. Our neural models do not suffer such bias as they are trained to predict one step ahead with the cross-entropy loss.

\begin{figure}
    \begin{center}
        \begin{subfigure}{0.32\textwidth}
            \includegraphics[width=\textwidth]{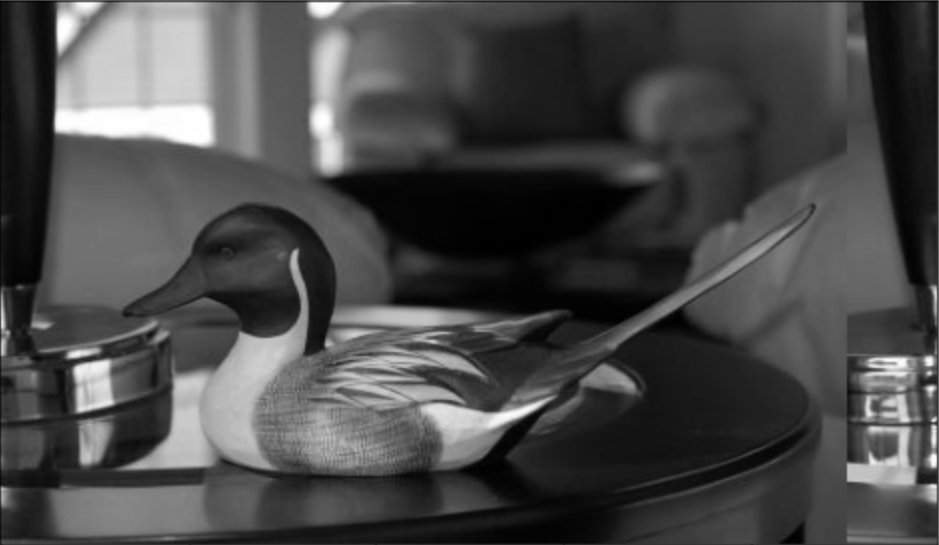}
            \caption{Original image}
        \end{subfigure}
        \hfill
        \begin{subfigure}{0.32\textwidth}
            \includegraphics[width=\textwidth]{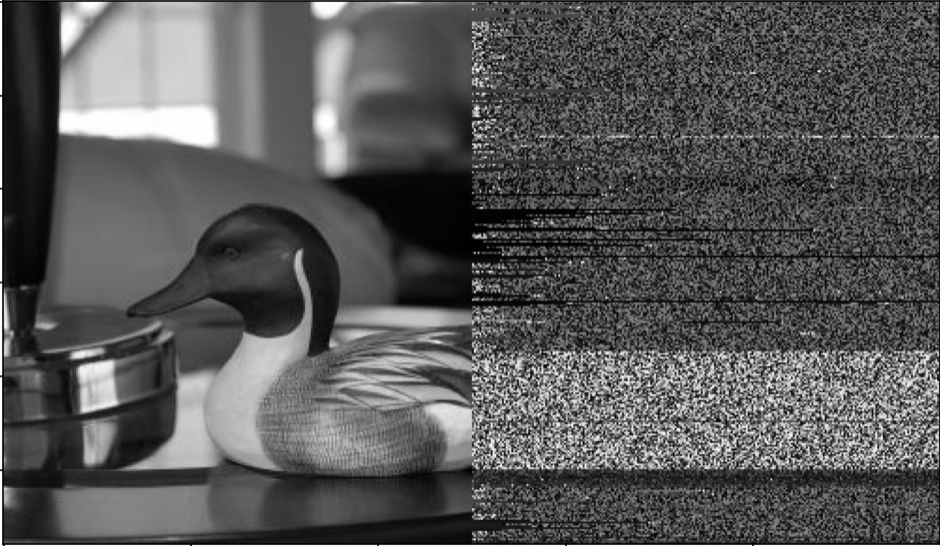}
            \caption{gzip (row-wise)}
        \end{subfigure}
        \hfill
        \begin{subfigure}{0.32\textwidth}
            \includegraphics[width=\textwidth]{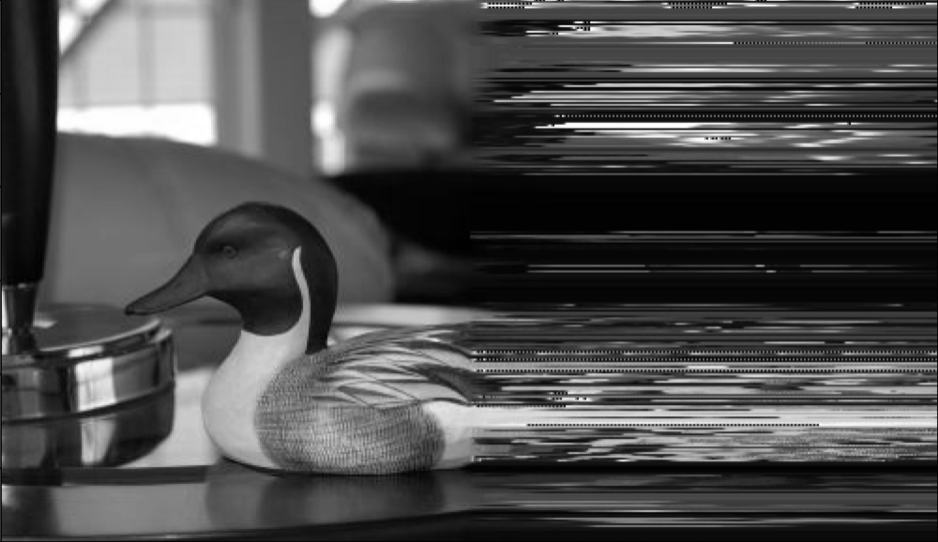}
            \caption{Chinchilla (row-wise)}
        \end{subfigure}
    \end{center}
    \caption{
        Compression-based generation for image data. 
        We condition gzip and Chinchilla on the first half of every row of the ImageNet image and then sample the remaining half autoregressively.
        Both models produce incoherent samples, but Chinchilla looks much less noisy than gzip.
    }
    \label{fig:generation-images}
\end{figure}

We compare the generative capabilities of gzip and Chinchilla 70B on images in \cref{fig:generation-images}.
Each image is a sampled from ImageNet with height 290 and width 500.
For each row in the image, we condition the model on the first 250 pixels and autoregressively generate the remaining 250 pixels, treating different rows as independent of each other (an oversimplification w.r.t. natural image statistics). 
We use the same byte conversions and tokenization details as explained in \cref{sec:experimental-details}.
Chinchilla 70B shows signatures of visually appropriate continuations (judged qualitatively), which tend to degrade with increased sample length as more and more error accumulates.
gzip produces much noisier completions.
We compare the generative performance of gzip and Chinchilla (1B, 7B, and 70B) across all three data modalities in \cref{fig:generation-text,fig:generation-images-extra,fig:generation-audio} for text, image, and audio data, respectively.

\subsection{Sequential Evolution of In-Context Compression}
\label{sec:experiments:in-context}

\begin{figure}
    \begin{center}
        \begin{subfigure}{0.32\textwidth}
            \includegraphics[width=\textwidth]{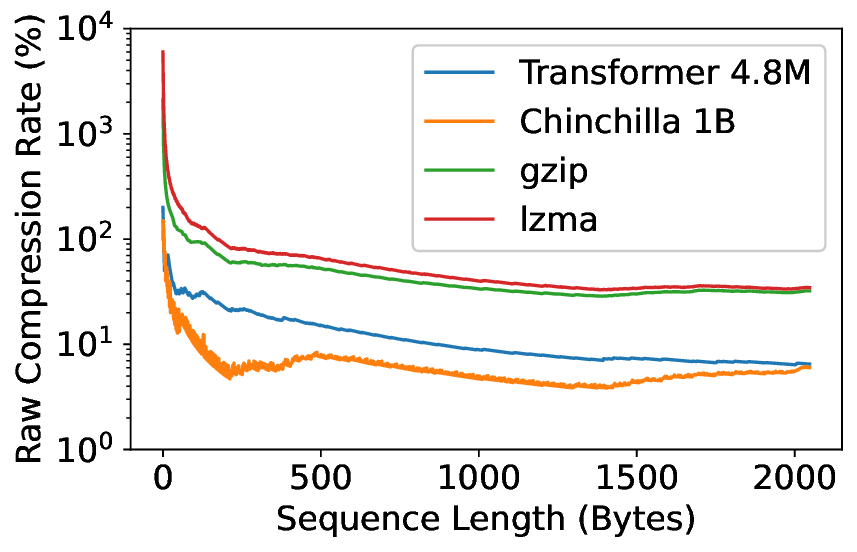}
            \caption{enwik9}
        \end{subfigure}
        \hfill
        \begin{subfigure}{0.32\textwidth}
            \includegraphics[width=\textwidth]{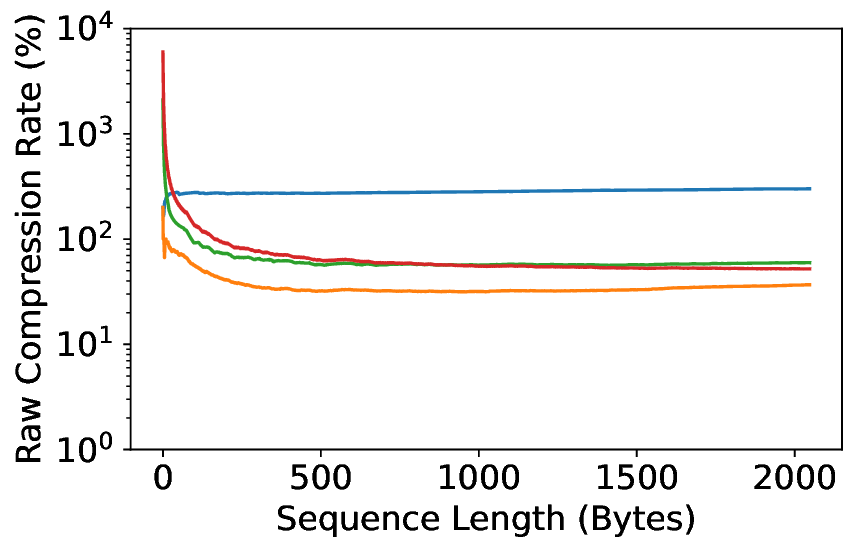}
            \caption{ImageNet}
        \end{subfigure}
        \hfill
        \begin{subfigure}{0.32\textwidth}
            \includegraphics[width=\textwidth]{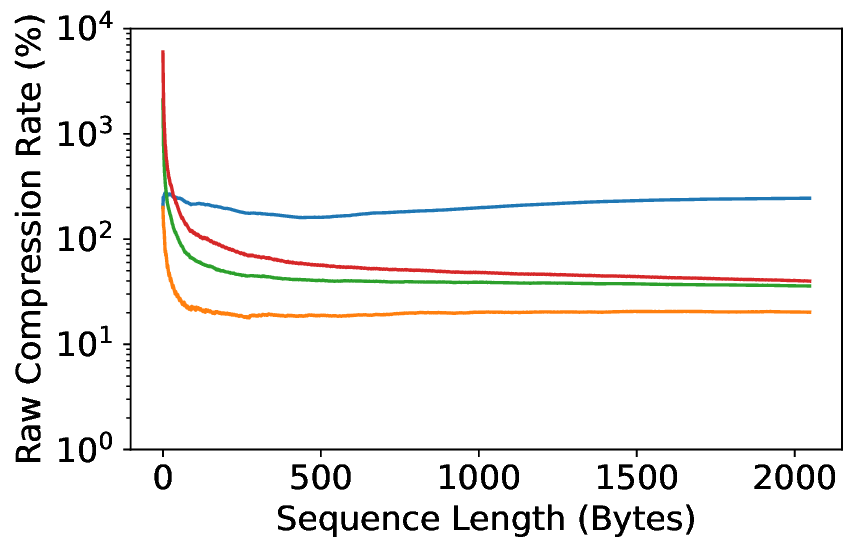}
            \caption{LibriSpeech}
        \end{subfigure}
    \end{center}
    \caption{
        In-context compression rate over sequence length.
        For every dataset, we compute the compression rate for all subsequences of 2048 bytes, averaged over 100 sequences.
    }
    \label{fig:incontext}
\end{figure}

Language models take a very different ``approach'' to compression compared to classical compressors.
Classical compressors have a small program size and optimize for a large context length to exploit sequential dependencies in the data.
In contrast, foundation models consist of billions of parameters, which enable rapid adaptation in their (relatively) short context window~\citep{genewein2023memory}.
Thus, arithmetic coding-based compressors rely heavily on the predictive models' in-context learning capabilities to achieve competitive compression performance.
We investigate this phenomenon in \cref{fig:incontext}, which visualizes the compression rate across sequence lengths for gzip, Chinchilla 1B and a Transformer pretrained on enwik8.
Intuitively, the longer the sequence, the more data the model can process in its context, and therefore, the better the compression.
As expected, most compression rates decrease quickly with increasing sequence length, indicating that the models learn some data statistics in-context, without any gradient-based training.
As in \cref{tab:compression-rates}, the Chinchilla model achieves the best compression rates across all three data modalities and sequence lengths.

\subsection{Tokenization Is Compression}
\label{sec:experiments:tokenization}

Transformers are generally not trained on raw input data but on tokenized versions thereof, both for efficiency and performance reasons. 
As a consequence, Transformers are trained on compressed data, with tokenizers acting as the compressor.
Since tokenization is known to have an impact on the generalization performance~\citep{radford2019language}, we investigate its impact on the compression rate in \cref{tab:tokenizers}.
Concretely, we train Transformers on enwik8 using different tokenizers: ASCII, \ie an alphabet of size 256 (no tokenization), and byte-pair encoding trained on enwik8, with various vocabulary sizes (1K, 2K, 5K, 10K, and 20K tokens).
Note that the tokenizations are lossless.

Increasing the number of tokens (\ie the ``alphabet size'') reduces the length of the sequence and thus increases the amount of information in a models context. 
However, decreasing the sequence length comes at a price: the number of tokens is larger, which makes the prediction task more challenging since reducing the entropy of the conditional distribution $\rho(x_i \mid x_{<i})$ is increasingly difficult for larger alphabet size.
In theory, as the tokenization is a lossless compression, the two effects should compensate.
In practice, we observe that if the model is small, increasing the number of possible tokens boosts the compression performance.
In contrast, for bigger models, it seems that the converse happens: having a larger token vocabulary harms the final compression rate of the model.
Nevertheless, short sequence lengths also help Transformers since their time complexity scales quadratically with context length, and it has been shown they do not generalize well to long contexts~\citep{deletang2023neural, ruoss2023randomized}.

\section{Related work}\label{sec:related-work}

\paragraph{Prediction \versus Compression}

Leveraging Shannon's source coding theorem~\citep{shannon1948mathematical}, a plethora of approaches exploit the connection between prediction and compression.
For example, context-tree weighting (CTW)~\citep{willems1995context} mixes the predictions of many underlying Markov models to achieve lossless compression via arithmetic coding~\citep{pasco1977source, rissanen1976generalized}.
Similarly, prediction by partial matching (PPM)~\citep{cleary1984data} also leverages arithmetic coding, but uses a contiguous context matching method to create  probability distributions based on the history of characters in a sequence. 
Likewise, PAQ8~\citep{knoll2012machine} uses a weighted combination of predictions from a large number of models (most of them based on context matching, but unlike PPM also noncontiguous context matches).
In a different setting, \citet{veness2015compress} demonstrated how to employ compression to obtain value estimates of a policy in an environment. 
\citet{frank2000text} and later \citet{teahan2003using} introduced the idea of classification with compressors. 
Recently, \citet{jiang2023low} applied this technique with NLP tasks, paired with a k-nearest-neighbour algorithm, in which gzip achieves good results.
\citet{jiang2022few} exploit the same idea but train the compressor on a vast amount of unlabeled data first. Further comparisons on the same tasks have been done between gzip and simple bag-of-words models~\citep{opitz2023gzip}.
Finally, \citet{oord2014student} apply arithmetic coding to image compression using Student distribution mixtures and Gaussian processes as predictors.

\paragraph{Compression With Neural Networks}

Prior work demonstrated that neural predictive distributions can be employed to perform lossless compression via arithmetic coding~\citep{schmidhuber1994predictive, schmidhuber1996sequential, mahoney2000fast, knoll2014cmix, cox2016syntactically, schiopu2018cnn, goyal2019deepzip, liu2019decmac, mentzer2019practical, mentzer2020learning, schiopu2020deep, rhee2022lc, mikolov2012statistical}.
Similarly, neural networks were also shown to achieve strong lossless compression rates when replacing arithmetic coding with asymmetric numeral systems~\citep{hoogeboom2019integer, kingma2019bit, townsend2019practical, barzen2022accelerated}.
While these approaches assume the existence of a separate training set, a different line of work investigated arithmetic coding-based neural compression in a purely online fashion, \ie training the model only on the data stream that is to be compressed~\citep{bellard2019lossless, goyal2020dzip, bellard2021nncp, mao2022trace}.
Concurrent work~\citep{valmeekam2023llmzip} also investigated lossless offline compression with foundation models, using arithmetic coding with Llama 2 (7B)~\citep{touvron2023llama2}.

\paragraph{Compression Biases: Tokenization, Model Size, etc.}

Much effort has been devoted to understanding the inductive biases of neural networks, in particular with respect to Natural Language Processing (NLP) and Transformers.
\citet{kudo2018sentencepiece} developed a tokenizer for NLP that improves over well-known techniques such as byte-pair encoding (BPE)~\citep{sennrich2016neural}, BPE dropout~\citep{provilkov2020bpedropout}, and subword regularization~\citep{kudo2018subword}.
In this paper, we show how these tokenization techniques act as pre-compressors for the data and can significantly affect the final compression rates when paired with a neural model.
Other work investigated different aspects of generalization~\citep{neyshabur2017exploring, mirchandani2023large, ge2023incontext}, which is equivalent to compression when accounting for the parameters' codejlength.
Finally, \citet{cheng2017survey} investigated compressing the neural models' parameters to further reduce the code length.

\section{Conclusion}\label{sec:conclusion}

In this paper we investigated how and why sequence modeling is equivalent to compression.
Arithmetic coding transforms a sequence model into a compressor, and, conversely, a compressor can be transformed into a predictor using its coding lengths to construct probability distributions following Shannon's entropy principle.
We evaluated large language models as compressors against various standard compressors and showed that they are competitive not only on text but also on modalities they have never been trained on (image and audio data).
We showed that the compression view provides novel insights on scaling laws since it takes the model size into account, unlike the log-loss objective, which is standard in current language modeling research.
Consequently, we showed that the optimal model size is inextricably linked to the dataset size and cannot be scaled without limit.

\subsection*{Acknowledgments}

We thank
Jörg Bornschein,
Nando de Freitas,
Slav Petrov, 
Zhengdong Wang,
and the anonymous reviewers for their helpful feedback and insightful discussions.

\bibliography{references}

\begin{thebibliography}{82}
\providecommand{\natexlab}[1]{#1}
\providecommand{\url}[1]{\texttt{#1}}
\expandafter\ifx\csname urlstyle\endcsname\relax
  \providecommand{\doi}[1]{doi: #1}\else
  \providecommand{\doi}{doi: \begingroup \urlstyle{rm}\Url}\fi

\bibitem[Barzen et~al.(2022)Barzen, Glazov, Geistert, and
  Sikora]{barzen2022accelerated}
Benjamin Lukas~Cajus Barzen, Fedor Glazov, Jonas Geistert, and Thomas Sikora.
\newblock Accelerated deep lossless image coding with unified paralleleized
  {GPU} coding architecture.
\newblock In \emph{{PCS}}, 2022.

\bibitem[Bellard(2019)]{bellard2019lossless}
Fabrice Bellard.
\newblock Lossless data compression with neural networks.
\newblock Technical report, Amarisoft, 2019.

\bibitem[Bellard(2021)]{bellard2021nncp}
Fabrice Bellard.
\newblock {NNCP} v2: Lossless data compression with transformer.
\newblock Technical report, Amarisoft, 2021.

\bibitem[Blier \& Ollivier(2018)Blier and Ollivier]{blier2018description}
L{\'{e}}onard Blier and Yann Ollivier.
\newblock The description length of deep learning models.
\newblock In \emph{NeurIPS}, 2018.

\bibitem[Bommasani et~al.(2021)]{bommasani2021opportunities}
Rishi Bommasani et~al.
\newblock On the opportunities and risks of foundation models.
\newblock \emph{arXiv:2108.07258}, 2021.

\bibitem[Boutell(1997)]{boutell1997png}
Thomas Boutell.
\newblock {PNG} (portable network graphics) specification version 1.0.
\newblock \emph{{RFC}}, 1997.

\bibitem[Brown et~al.(2020)Brown, Ryder, Subbiah, et~al.]{brown2020language}
Tom~B. Brown, Benjamin Mannand~Nick Ryder, Melanie Subbiah, et~al.
\newblock Language models are few-shot learners.
\newblock In \emph{NeurIPS}, 2020.

\bibitem[Bubeck et~al.(2023)Bubeck, Chandrasekaran, Eldan, Gehrke, Horvitz,
  Kamar, Lee, Lee, Li, Lundberg, Nori, Palangi, Ribeiro, and
  Zhang]{bubeck2023sparks}
S{\'{e}}bastien Bubeck, Varun Chandrasekaran, Ronen Eldan, Johannes Gehrke,
  Eric Horvitz, Ece Kamar, Peter Lee, Yin~Tat Lee, Yuanzhi Li, Scott~M.
  Lundberg, Harsha Nori, Hamid Palangi, Marco~T{\'{u}}lio Ribeiro, and
  Yi~Zhang.
\newblock Sparks of artificial general intelligence: Early experiments with
  {GPT-4}.
\newblock \emph{arXiv:2303.12712}, 2023.

\bibitem[Bulatov et~al.(2023)Bulatov, Kuratov, and Burtsev]{bulatov2023scaling}
Aydar Bulatov, Yuri Kuratov, and Mikhail~S. Burtsev.
\newblock Scaling transformer to 1m tokens and beyond with {RMT}.
\newblock \emph{arXiv:2304.11062}, 2023.

\bibitem[Cheng et~al.(2017)Cheng, Wang, Zhou, and Zhang]{cheng2017survey}
Yu~Cheng, Duo Wang, Pan Zhou, and Tao Zhang.
\newblock A survey of model compression and acceleration for deep neural
  networks.
\newblock \emph{arXiv:1710.09282}, 2017.

\bibitem[Cleary \& Witten(1984)Cleary and Witten]{cleary1984data}
John~G. Cleary and Ian~H. Witten.
\newblock Data compression using adaptive coding and partial string matching.
\newblock \emph{{IEEE} Trans. Commun.}, 1984.

\bibitem[Coalson(2008)]{coalson2008flac}
Josh Coalson.
\newblock Free lossless audio codec, 2008.
\newblock URL \url{https://xiph.org/flac}.

\bibitem[Cox(2016)]{cox2016syntactically}
David Cox.
\newblock Syntactically informed text compression with recurrent neural
  networks.
\newblock \emph{arXiv:1608.02893}, 2016.

\bibitem[Del{\'{e}}tang et~al.(2023)Del{\'{e}}tang, Ruoss, Grau{-}Moya,
  Genewein, Wenliang, Catt, Cundy, Hutter, Legg, Veness, and
  Ortega]{deletang2023neural}
Gr{\'{e}}goire Del{\'{e}}tang, Anian Ruoss, Jordi Grau{-}Moya, Tim Genewein,
  Li~Kevin Wenliang, Elliot Catt, Chris Cundy, Marcus Hutter, Shane Legg, Joel
  Veness, and Pedro~A. Ortega.
\newblock Neural networks and the chomsky hierarchy.
\newblock In \emph{{ICLR}}, 2023.

\bibitem[Deutsch(1996)]{deutsch1996gzip}
Peter Deutsch.
\newblock {GZIP} file format specification version 4.3.
\newblock \emph{{RFC}}, 1996.

\bibitem[Duda(2009)]{duda2009asymmetric}
Jarek Duda.
\newblock Asymmetric numeral systems.
\newblock \emph{arXiv:0902.0271}, 2009.

\bibitem[Frank et~al.(2000)Frank, Chui, and Witten]{frank2000text}
Eibe Frank, Chang Chui, and Ian~H. Witten.
\newblock Text categorization using compression models.
\newblock In \emph{Data Compression Conference}, 2000.

\bibitem[Ge et~al.(2023)Ge, Hu, Wang, Chen, and Wei]{ge2023incontext}
Tao Ge, Jing Hu, Xun Wang, Si{-}Qing Chen, and Furu Wei.
\newblock In-context autoencoder for context compression in a large language
  model.
\newblock \emph{arXiv:2307.06945}, 2023.

\bibitem[Genewein et~al.(2023)Genewein, Del{\'{e}}tang, Ruoss, Wenliang, Catt,
  Dutordoir, Grau{-}Moya, Orseau, Hutter, and Veness]{genewein2023memory}
Tim Genewein, Gr{\'{e}}goire Del{\'{e}}tang, Anian Ruoss, Li~Kevin Wenliang,
  Elliot Catt, Vincent Dutordoir, Jordi Grau{-}Moya, Laurent Orseau, Marcus
  Hutter, and Joel Veness.
\newblock Memory-based meta-learning on non-stationary distributions.
\newblock In \emph{{ICML}}, 2023.

\bibitem[Goyal et~al.(2019)Goyal, Tatwawadi, Chandak, and
  Ochoa]{goyal2019deepzip}
Mohit Goyal, Kedar Tatwawadi, Shubham Chandak, and Idoia Ochoa.
\newblock Deepzip: Lossless data compression using recurrent neural networks.
\newblock In \emph{{DCC}}, 2019.

\bibitem[Goyal et~al.(2020)Goyal, Tatwawadi, Chandak, and Ochoa]{goyal2020dzip}
Mohit Goyal, Kedar Tatwawadi, Shubham Chandak, and Idoia Ochoa.
\newblock Dzip: Improved general-purpose lossless compression based on novel
  neural network modeling.
\newblock In \emph{{DCC}}, 2020.

\bibitem[Guo et~al.(2022)Guo, Ainslie, Uthus, Onta{\~{n}}{\'{o}}n, Ni, Sung,
  and Yang]{guo2022longt5}
Mandy Guo, Joshua Ainslie, David~C. Uthus, Santiago Onta{\~{n}}{\'{o}}n, Jianmo
  Ni, Yun{-}Hsuan Sung, and Yinfei Yang.
\newblock Longt5: Efficient text-to-text transformer for long sequences.
\newblock In \emph{{NAACL-HLT} (Findings)}, 2022.

\bibitem[Hoffmann et~al.(2022)Hoffmann, Borgeaud, Mensch,
  et~al.]{hoffmann2022training}
Jordan Hoffmann, Sebastian Borgeaud, Arthur Mensch, et~al.
\newblock Training compute-optimal large language models.
\newblock \emph{arXiv:2203.15556}, 2022.

\bibitem[Hoogeboom et~al.(2019)Hoogeboom, Peters, van~den Berg, and
  Welling]{hoogeboom2019integer}
Emiel Hoogeboom, Jorn W.~T. Peters, Rianne van~den Berg, and Max Welling.
\newblock Integer discrete flows and lossless compression.
\newblock In \emph{NeurIPS}, 2019.

\bibitem[Howard \& Vitter(1991)Howard and Vitter]{howard1991analysis}
Paul~G. Howard and Jeffrey~Scott Vitter.
\newblock Analysis of arithmetic coding for data compression.
\newblock In \emph{Data Compression Conference}, 1991.

\bibitem[Huffman(1952)]{huffman1952method}
David~A. Huffman.
\newblock A method for the construction of minimum-redundancy codes.
\newblock \emph{Proceedings of the IRE}, 1952.

\bibitem[Hutter(2005)]{hutter2005universal}
Marcus Hutter.
\newblock \emph{Universal Artificial Intellegence - Sequential Decisions Based
  on Algorithmic Probability}.
\newblock Springer, 2005.

\bibitem[Hutter(2006)]{hutter2006prize}
Marcus Hutter.
\newblock 500'000\texteuro~prize for compressing human knowledge, 2006.
\newblock URL \url{http://prize.hutter1.net}.

\bibitem[Jiang et~al.(2022)Jiang, Dai, Xin, Li, and Lin]{jiang2022few}
Zhiying Jiang, Yiqin Dai, Ji~Xin, Ming Li, and Jimmy Lin.
\newblock Few-shot non-parametric learning with deep latent variable model.
\newblock In \emph{NeurIPS}, 2022.

\bibitem[Jiang et~al.(2023)Jiang, Yang, Tsirlin, Tang, Dai, and
  Lin]{jiang2023low}
Zhiying Jiang, Matthew Y.~R. Yang, Mikhail Tsirlin, Raphael Tang, Yiqin Dai,
  and Jimmy Lin.
\newblock "low-resource" text classification: {A} parameter-free classification
  method with compressors.
\newblock In \emph{{ACL} (Findings)}, 2023.

\bibitem[Kaplan et~al.(2020)Kaplan, McCandlish, Henighan, Brown, Chess, Child,
  Gray, Radford, Wu, and Amodei]{kaplan2020scaling}
Jared Kaplan, Sam McCandlish, Tom Henighan, Tom~B. Brown, Benjamin Chess, Rewon
  Child, Scott Gray, Alec Radford, Jeffrey Wu, and Dario Amodei.
\newblock Scaling laws for neural language models.
\newblock \emph{arXiv:2001.08361}, 2020.

\bibitem[Kingma et~al.(2019)Kingma, Abbeel, and Ho]{kingma2019bit}
Friso~H. Kingma, Pieter Abbeel, and Jonathan Ho.
\newblock Bit-swap: Recursive bits-back coding for lossless compression with
  hierarchical latent variables.
\newblock In \emph{{ICML}}, 2019.

\bibitem[Knoll(2014)]{knoll2014cmix}
Byron Knoll.
\newblock {CMIX}, 2014.
\newblock URL \url{http://www.byronknoll.com/cmix.html}.

\bibitem[Knoll \& de~Freitas(2012)Knoll and de~Freitas]{knoll2012machine}
Byron Knoll and Nando de~Freitas.
\newblock A machine learning perspective on predictive coding with {PAQ8}.
\newblock In \emph{{DCC}}, 2012.

\bibitem[Kolmogorov(1998)]{kolmogorov1998table}
Andrei~N. Kolmogorov.
\newblock On tables of random numbers.
\newblock \emph{Theoretical Computer Science}, 1998.

\bibitem[Kudo(2018)]{kudo2018subword}
Taku Kudo.
\newblock Subword regularization: Improving neural network translation models
  with multiple subword candidates.
\newblock In \emph{{ACL} {(1)}}, 2018.

\bibitem[Kudo \& Richardson(2018)Kudo and Richardson]{kudo2018sentencepiece}
Taku Kudo and John Richardson.
\newblock Sentencepiece: {A} simple and language independent subword tokenizer
  and detokenizer for neural text processing.
\newblock In \emph{{EMNLP} (Demonstration)}, 2018.

\bibitem[Laskin et~al.(2023)Laskin, Wang, et~al.]{laskin2023incontext}
Michael Laskin, Luyu Wang, et~al.
\newblock In-context reinforcement learning with algorithm distillation.
\newblock In \emph{{ICLR}}. OpenReview.net, 2023.

\bibitem[Li \& Vit{\'{a}}nyi(2019)Li and Vit{\'{a}}nyi]{li2019introduction}
Ming Li and Paul M.~B. Vit{\'{a}}nyi.
\newblock \emph{An Introduction to Kolmogorov Complexity and Its Applications,
  4th Edition}.
\newblock Springer, 2019.

\bibitem[Liu et~al.(2019)Liu, Xu, and Li]{liu2019decmac}
Qian Liu, Yiling Xu, and Zhu Li.
\newblock {DecMac}: {A} deep context model for high efficiency arithmetic
  coding.
\newblock In \emph{{ICAIIC}}, 2019.

\bibitem[MacKay(2003)]{mackay2003information}
David J.~C. MacKay.
\newblock \emph{Information theory, inference, and learning algorithms}.
\newblock Cambridge University Press, 2003.

\bibitem[Mahoney(2000)]{mahoney2000fast}
Matthew~V. Mahoney.
\newblock Fast text compression with neural networks.
\newblock In \emph{{FLAIRS}}, 2000.

\bibitem[Mao et~al.(2022)Mao, Cui, Kuo, and Xue]{mao2022trace}
Yu~Mao, Yufei Cui, Tei{-}Wei Kuo, and Chun~Jason Xue.
\newblock {TRACE:} {A} fast transformer-based general-purpose lossless
  compressor.
\newblock In \emph{{WWW}}, 2022.

\bibitem[Mentzer et~al.(2019)Mentzer, Agustsson, Tschannen, Timofte, and
  Gool]{mentzer2019practical}
Fabian Mentzer, Eirikur Agustsson, Michael Tschannen, Radu Timofte, and Luc~Van
  Gool.
\newblock Practical full resolution learned lossless image compression.
\newblock In \emph{{CVPR}}, 2019.

\bibitem[Mentzer et~al.(2020)Mentzer, Gool, and Tschannen]{mentzer2020learning}
Fabian Mentzer, Luc~Van Gool, and Michael Tschannen.
\newblock Learning better lossless compression using lossy compression.
\newblock In \emph{{CVPR}}, 2020.

\bibitem[Mikolov(2012)]{mikolov2012statistical}
Tomas Mikolov.
\newblock \emph{Statistical Language Models Based on Neural Networks}.
\newblock PhD thesis, Brno Universtiy of Technology, 2012.

\bibitem[Mirchandani et~al.(2023)Mirchandani, Xia, Florence, Ichter, Driess,
  Arenas, Rao, Sadigh, and Zeng]{mirchandani2023large}
Suvir Mirchandani, Fei Xia, Pete Florence, Brian Ichter, Danny Driess,
  Montserrat~Gonzalez Arenas, Kanishka Rao, Dorsa Sadigh, and Andy Zeng.
\newblock Large language models as general pattern machines.
\newblock \emph{arXiv:2307.04721}, 2023.

\bibitem[Neyshabur et~al.(2017)Neyshabur, Bhojanapalli, McAllester, and
  Srebro]{neyshabur2017exploring}
Behnam Neyshabur, Srinadh Bhojanapalli, David McAllester, and Nati Srebro.
\newblock Exploring generalization in deep learning.
\newblock In \emph{{NIPS}}, 2017.

\bibitem[Opitz(2023)]{opitz2023gzip}
Juri Opitz.
\newblock Gzip versus bag-of-words for text classification, 2023.

\bibitem[Panayotov et~al.(2015)Panayotov, Chen, Povey, and
  Khudanpur]{panayotov2015librispeech}
Vassil Panayotov, Guoguo Chen, Daniel Povey, and Sanjeev Khudanpur.
\newblock Librispeech: An {ASR} corpus based on public domain audio books.
\newblock In \emph{{ICASSP}}, 2015.

\bibitem[Pasco(1977)]{pasco1977source}
Richard~C. Pasco.
\newblock Source coding algorithms for fast data compression (ph.d. thesis
  abstr.).
\newblock \emph{{IEEE} Trans. Inf. Theory}, 1977.

\bibitem[Pavlov(2019)]{pavlov20197z}
Igor Pavlov.
\newblock {7z Format}, 2019.
\newblock URL \url{http://www.7-zip.org/7z.html}.

\bibitem[Provilkov et~al.(2020)Provilkov, Emelianenko, and
  Voita]{provilkov2020bpedropout}
Ivan Provilkov, Dmitrii Emelianenko, and Elena Voita.
\newblock Bpe-dropout: Simple and effective subword regularization.
\newblock In \emph{{ACL}}, 2020.

\bibitem[Radford et~al.(2019)Radford, Wu, Child, Luan, Amodei, and
  Sutskever]{radford2019language}
Alec Radford, Jeff Wu, Rewon Child, David Luan, Dario Amodei, and Ilya
  Sutskever.
\newblock Language models are unsupervised multitask learners.
\newblock Technical report, Open{AI}, 2019.

\bibitem[Rae et~al.(2021)]{rae2021scaling}
Jack~W. Rae et~al.
\newblock Scaling language models: Methods, analysis {\&} insights from
  training gopher.
\newblock \emph{arXiv:2112.11446}, 2021.

\bibitem[Rathmanner \& Hutter(2011)Rathmanner and
  Hutter]{rathmanner2011philosophical}
Samuel Rathmanner and Marcus Hutter.
\newblock A philosophical treatise of universal induction.
\newblock \emph{Entropy}, 2011.

\bibitem[Rhee et~al.(2022)Rhee, Jang, Kim, and Cho]{rhee2022lc}
Hochang Rhee, Yeong~Il Jang, Seyun Kim, and Nam~Ik Cho.
\newblock {LC-FDNet}: Learned lossless image compression with frequency
  decomposition network.
\newblock In \emph{{CVPR}}, 2022.

\bibitem[Rissanen(1976)]{rissanen1976generalized}
Jorma Rissanen.
\newblock Generalized kraft inequality and arithmetic coding.
\newblock \emph{{IBM} J. Res. Dev.}, 1976.

\bibitem[Ruoss et~al.(2023)Ruoss, Del{\'{e}}tang, Genewein, Grau{-}Moya,
  Csord{\'{a}}s, Bennani, Legg, and Veness]{ruoss2023randomized}
Anian Ruoss, Gr{\'{e}}goire Del{\'{e}}tang, Tim Genewein, Jordi Grau{-}Moya,
  R{\'{o}}bert Csord{\'{a}}s, Mehdi Bennani, Shane Legg, and Joel Veness.
\newblock Randomized positional encodings boost length generalization of
  transformers.
\newblock In \emph{{ACL} {(2)}}, 2023.

\bibitem[Russakovsky et~al.(2015)Russakovsky, Deng, Su, Krause, Satheesh, Ma,
  Huang, Karpathy, Khosla, Bernstein, Berg, and
  Fei{-}Fei]{russakovsky2015imagenet}
Olga Russakovsky, Jia Deng, Hao Su, Jonathan Krause, Sanjeev Satheesh, Sean Ma,
  Zhiheng Huang, Andrej Karpathy, Aditya Khosla, Michael~S. Bernstein,
  Alexander~C. Berg, and Li~Fei{-}Fei.
\newblock Imagenet large scale visual recognition challenge.
\newblock \emph{Int. J. Comput. Vis.}, 2015.

\bibitem[Schiopu \& Munteanu(2020)Schiopu and Munteanu]{schiopu2020deep}
Ionut Schiopu and Adrian Munteanu.
\newblock Deep-learning-based lossless image coding.
\newblock \emph{{IEEE} Trans. Circuits Syst. Video Technol.}, 2020.

\bibitem[Schiopu et~al.(2018)Schiopu, Liu, and Munteanu]{schiopu2018cnn}
Ionut Schiopu, Yu~Liu, and Adrian Munteanu.
\newblock {CNN}-based prediction for lossless coding of photographic images.
\newblock In \emph{{PCS}}, 2018.

\bibitem[Schmidhuber \& Heil(1994)Schmidhuber and
  Heil]{schmidhuber1994predictive}
J{\"{u}}rgen Schmidhuber and Stefan Heil.
\newblock Predictive coding with neural nets: Application to text compression.
\newblock In \emph{{NIPS}}, pp.\  1047--1054. {MIT} Press, 1994.

\bibitem[Schmidhuber \& Heil(1996)Schmidhuber and
  Heil]{schmidhuber1996sequential}
J{\"{u}}rgen Schmidhuber and Stefan Heil.
\newblock Sequential neural text compression.
\newblock \emph{{IEEE} Trans. Neural Networks}, 1996.

\bibitem[Sennrich et~al.(2016)Sennrich, Haddow, and Birch]{sennrich2016neural}
Rico Sennrich, Barry Haddow, and Alexandra Birch.
\newblock Neural machine translation of rare words with subword units.
\newblock In \emph{{ACL} {(1)}}, 2016.

\bibitem[Shannon(1948)]{shannon1948mathematical}
Claude~E. Shannon.
\newblock A mathematical theory of communication.
\newblock \emph{Bell Syst. Tech. J.}, 1948.

\bibitem[Solomonoff(1964{\natexlab{a}})]{solomonoff1964formal1}
Ray~J. Solomonoff.
\newblock A formal theory of inductive inference. part {I}.
\newblock \emph{Inf. Control.}, 1964{\natexlab{a}}.

\bibitem[Solomonoff(1964{\natexlab{b}})]{solomonoff1964formal2}
Ray~J. Solomonoff.
\newblock A formal theory of inductive inference. part {II}.
\newblock \emph{Inf. Control.}, 1964{\natexlab{b}}.

\bibitem[Tao et~al.(2022)Tao, Hou, Zhang, Shang, Jiang, Liu, Luo, and
  Wong]{tao2022compression}
Chaofan Tao, Lu~Hou, Wei Zhang, Lifeng Shang, Xin Jiang, Qun Liu, Ping Luo, and
  Ngai Wong.
\newblock Compression of generative pre-trained language models via
  quantization.
\newblock In \emph{{ACL} {(1)}}, 2022.

\bibitem[Teahan \& Harper(2003)Teahan and Harper]{teahan2003using}
William~J. Teahan and David~J. Harper.
\newblock \emph{Using Compression-Based Language Models for Text
  Categorization}, pp.\  141--165.
\newblock Springer Netherlands, 2003.

\bibitem[Touvron et~al.(2023{\natexlab{a}})Touvron, Lavril, Izacard,
  et~al.]{touvron2023llama}
Hugo Touvron, Thibaut Lavril, Gautier Izacard, et~al.
\newblock Llama: Open and efficient foundation language models.
\newblock \emph{arXiv:2302.13971}, 2023{\natexlab{a}}.

\bibitem[Touvron et~al.(2023{\natexlab{b}})Touvron, Martin, Stone, Albert,
  Almahairi, Babaei, Bashlykov, Batra, Bhargava, Bhosale, Bikel, Blecher,
  Canton{-}Ferrer, Chen, Cucurull, Esiobu, Fernandes, Fu, Fu, Fuller, Gao,
  Goswami, Goyal, Hartshorn, Hosseini, Hou, Inan, Kardas, Kerkez, Khabsa,
  Kloumann, Korenev, Koura, Lachaux, Lavril, Lee, Liskovich, Lu, Mao, Martinet,
  Mihaylov, Mishra, Molybog, Nie, Poulton, Reizenstein, Rungta, Saladi,
  Schelten, Silva, Smith, Subramanian, Tan, Tang, Taylor, Williams, Kuan, Xu,
  Yan, Zarov, Zhang, Fan, Kambadur, Narang, Rodriguez, Stojnic, Edunov, and
  Scialom]{touvron2023llama2}
Hugo Touvron, Louis Martin, Kevin Stone, Peter Albert, Amjad Almahairi, Yasmine
  Babaei, Nikolay Bashlykov, Soumya Batra, Prajjwal Bhargava, Shruti Bhosale,
  Dan Bikel, Lukas Blecher, Cristian Canton{-}Ferrer, Moya Chen, Guillem
  Cucurull, David Esiobu, Jude Fernandes, Jeremy Fu, Wenyin Fu, Brian Fuller,
  Cynthia Gao, Vedanuj Goswami, Naman Goyal, Anthony Hartshorn, Saghar
  Hosseini, Rui Hou, Hakan Inan, Marcin Kardas, Viktor Kerkez, Madian Khabsa,
  Isabel Kloumann, Artem Korenev, Punit~Singh Koura, Marie{-}Anne Lachaux,
  Thibaut Lavril, Jenya Lee, Diana Liskovich, Yinghai Lu, Yuning Mao, Xavier
  Martinet, Todor Mihaylov, Pushkar Mishra, Igor Molybog, Yixin Nie, Andrew
  Poulton, Jeremy Reizenstein, Rashi Rungta, Kalyan Saladi, Alan Schelten, Ruan
  Silva, Eric~Michael Smith, Ranjan Subramanian, Xiaoqing~Ellen Tan, Binh Tang,
  Ross Taylor, Adina Williams, Jian~Xiang Kuan, Puxin Xu, Zheng Yan, Iliyan
  Zarov, Yuchen Zhang, Angela Fan, Melanie Kambadur, Sharan Narang,
  Aur{\'{e}}lien Rodriguez, Robert Stojnic, Sergey Edunov, and Thomas Scialom.
\newblock Llama 2: Open foundation and fine-tuned chat models.
\newblock \emph{arXiv:2307.09288}, 2023{\natexlab{b}}.

\bibitem[Townsend et~al.(2019)Townsend, Bird, and
  Barber]{townsend2019practical}
James Townsend, Thomas Bird, and David Barber.
\newblock Practical lossless compression with latent variables using bits back
  coding.
\newblock In \emph{{ICLR} (Poster)}, 2019.

\bibitem[Valmeekam et~al.(2023)Valmeekam, Narayanan, Kalathil, Chamberland, and
  Shakkottai]{valmeekam2023llmzip}
Chandra Shekhara~Kaushik Valmeekam, Krishna Narayanan, Dileep Kalathil,
  Jean{-}Fran{\c{c}}ois Chamberland, and Srinivas Shakkottai.
\newblock Llmzip: Lossless text compression using large language models.
\newblock \emph{arXiv:2306.04050}, 2023.

\bibitem[van~den Oord \& Schrauwen(2014)van~den Oord and
  Schrauwen]{oord2014student}
A{\"{a}}ron van~den Oord and Benjamin Schrauwen.
\newblock The student-t mixture as a natural image patch prior with application
  to image compression.
\newblock \emph{J. Mach. Learn. Res.}, 2014.

\bibitem[Vaswani et~al.(2017)Vaswani, Shazeer, Parmar, Uszkoreit, Jones, Gomez,
  Kaiser, and Polosukhin]{vaswani2017attention}
Ashish Vaswani, Noam Shazeer, Niki Parmar, Jakob Uszkoreit, Llion Jones,
  Aidan~N. Gomez, Lukasz Kaiser, and Illia Polosukhin.
\newblock Attention is all you need.
\newblock In \emph{{NIPS}}, 2017.

\bibitem[Veness et~al.(2015)Veness, Bellemare, Hutter, Chua, and
  Desjardins]{veness2015compress}
Joel Veness, Marc~G. Bellemare, Marcus Hutter, Alvin Chua, and Guillaume
  Desjardins.
\newblock Compress and control.
\newblock In \emph{{AAAI}}, 2015.

\bibitem[Wei et~al.(2022)Wei, Wang, Schuurmans, Bosma, Ichter, Xia, Chi, Le,
  and Zhou]{wei2022chain}
Jason Wei, Xuezhi Wang, Dale Schuurmans, Maarten Bosma, Brian Ichter, Fei Xia,
  Ed~H. Chi, Quoc~V. Le, and Denny Zhou.
\newblock Chain-of-thought prompting elicits reasoning in large language
  models.
\newblock In \emph{NeurIPS}, 2022.

\bibitem[Welch(1984)]{welch1984technique}
Terry~A. Welch.
\newblock A technique for high-performance data compression.
\newblock \emph{Computer}, 1984.

\bibitem[Willems et~al.(1995)Willems, Shtarkov, and
  Tjalkens]{willems1995context}
Frans M.~J. Willems, Yuri~M. Shtarkov, and Tjalling~J. Tjalkens.
\newblock The context-tree weighting method: basic properties.
\newblock \emph{{IEEE} Trans. Inf. Theory}, 1995.

\bibitem[Witten et~al.(1987)Witten, Neal, and Cleary]{witten1987arithmetic}
Ian~H. Witten, Radford~M. Neal, and John~G. Cleary.
\newblock Arithmetic coding for data compression.
\newblock \emph{Commun. {ACM}}, 1987.

\bibitem[Zaheer et~al.(2020)Zaheer, Guruganesh, Dubey, Ainslie, Alberti,
  Onta{\~{n}}{\'{o}}n, Pham, Ravula, Wang, Yang, and Ahmed]{zaheer2020big}
Manzil Zaheer, Guru Guruganesh, Kumar~Avinava Dubey, Joshua Ainslie, Chris
  Alberti, Santiago Onta{\~{n}}{\'{o}}n, Philip Pham, Anirudh Ravula, Qifan
  Wang, Li~Yang, and Amr Ahmed.
\newblock Big bird: Transformers for longer sequences.
\newblock In \emph{NeurIPS}, 2020.

\end{thebibliography}
\bibliographystyle{iclr2024_conference}

\clearpage

\appendix
\counterwithin{figure}{section}
\counterwithin{table}{section}

\section{Arithmetic Coding}
\label{sec:arithmetic-coding}

Here we provide a step-by-step explanation of the arithmetic encoding example visualized in \cref{fig:overview}.
Recall from \cref{sec:background} that arithmetic encoding iteratively partitions the interval $I = [0, 1)$ according to a predictive model $P$ and an input string, \ie $AIXI$ for \cref{fig:overview}.

First, we construct the intervals for the first token, corresponding to $P(\cdot)$:
\begin{itemize}
    \item $[0, 0.45)$ for $P(A) = 0.45$
    \item $[0.45, 0.75)$ for $P(I) = 0.3$
    \item $[0.75, 1)$ for $P(X) = 0.25$
\end{itemize}
Since the first token is $A$, we set $I = [0, 0.45)$ and iterate. Thus, the intervals for $P(\cdot | A)$ are:
\begin{itemize}
    \item $[0.2 * 0, 0.2 * 0.45) = [0, 0.09)$ for $P(A | A) = 0.2$
    \item $[0.09, (0.2 + 0.6) * 0.45) = [0.09, 0.36)$ for $P(I | A) = 0.3$
    \item $[0.36, (0.2 + 0.6 + 0.2) * 0.45) = [0.36, 0.45)$ for $P(X | A) = 0.2$
\end{itemize}
Since the next token is $I$, we set $I = [0, 0.45)$ and so on. We terminate with $I = [0.322, 0341)$ for $AIXI$. Next, arithmetic will compute the binary sequence corresponding to iteratively splitting the interval $[0, 1)$ in half until it is fully contained in $I$. Concretely, this yields the binary sequence.
\begin{itemize}
    \item $b0       \rightarrow [0, 0.5)$
    \item $b01      \rightarrow [0.25, 0.5)$ 
    \item $b010     \rightarrow [0.25, 0.375)$
    \item $b0101    \rightarrow [0.3125, 0.375)$
    \item $b01010   \rightarrow [0.3125, 0.34375)$
    \item $b010101  \rightarrow [0.328125, 0.34375)$
    \item $b0101010 \rightarrow [0.328125, 0.3359375)$
\end{itemize}
As $[0.328125, 0.3359375)$ is fully contained in $I = [0.322, 0341)$, the compressed output is $0101010$, which consists of 7 bits as opposed to the 4 bytes used to encode $AIXI$.

\section{Experimental Details}
\label{sec:experimental-details}

\subsection{Data manipulation}

\paragraph{ASCII as a standard} The neural models we use (Transformers trained specifically on enwik, and Chinchilla which is trained on a vast and diverse set of text from various sources) take text as input. In theory, they could take in any Unicode character, but unfortunately not all bytes in the range [128, 256]. However, any ASCII character, i.e. in the range [0, 127] is a valid input. Therefore, we chose to compress all bytes sequences by first mapping them to the range [0, 127]. We account for the loss of 1 bit by updating the compression ratio accordingly, if necessary. Each character that needs some processing adds 1 bit to the final compressed sequence.

\paragraph{Compressing text} Text compression is very straightforward. As state above, the text is encoded in ASCII format. If the character is special, we zero the most significant bit, such that the value falls back in the range [0, 127], suitable for ASCII encoding. We append this lost bit at the end of the compressed byte sequence, to account for it in the compression rate.

\paragraph{Compressing images} Image data comes as patches of size (32, 64), each pixel being grayscaled and therefore encoded on 1 byte. This byte gives the brightness of the pixel, 0 being completely black and 255 being completely white. We then flatten this patch to get a sequence of 2048 bytes. Note that it means we lose some 2 dimension correlation between the pixels: the last pixel of the first line will be right next to the first pixel of the second line. For each byte, to map it into the range [0, 127], we simply divide it by 2, and lose the least significant bit. This is performed via simple byte shifting. Note that we do this transformation for every byte, and not only those in the range [128, 255], for consistency. We append the lost bit to the end of the compressed byte sequence, as explained above.

\paragraph{Compressing sound} Sound data from LibriSpeech is sampled at 16Khz, with a sample size of 2 bytes (int16). We manually reduce the sample size to 1 byte. We then split the raw dataset into chunks of size 2048 bytes each, which correspond to roughly 64 milliseconds of speech. We checked whether reducing the sample rate to have longer sequences had a significant impact on the relative compression powers of the models, and we concluded that it had not. We chose to keep the original rate for our experiments. Exactly as for images, for each byte, to map it into the range [0, 127], we simply divide it by 2, and lose the least significant bit. Note that we do this transformation for every byte, and not only those in the range [128, 255], for consistency. The lost bit is also appended at the end of the compressed byte sequence, as explained above.

\subsection{Large language models tokenization}

As described in the last subsection, the data fed to the large language models we use (Chinchilla and LLama2) is an ASCII string of exactly 2048 characters. However, the models immediately tokenizes the string using SentencePiece~\citep{kudo2018sentencepiece}. The string is transformed into a sequence of integer tokens between 0 and $T$, $T$ being the vocabulary size (they both use $T = 32000$). Note that the length of the sequence has now completely changed, and depends on the input: tokenization is already a form of lossless compression. This sequence is fed into the big pretrained Transformer model, which gives us the conditionals $\hat{\rho}(y | x_{<i})$ for all histories $x_{<i}$ and tokens in the alphabet $y$. Denoting the length of the sequence after tokenization as $l$, we obtain $l * T$ log-probabilities. We can pass them to an arithmetic encoder of vocabulary size $T$, to encode the sequence into bits. This is our final compressed sequence, which size in bytes is compared with the initial size, i.e., 2048 bytes.

In practice, the large models had only access to the top-k next token log-probabilities, for each context. We chose $k=100$, which almost fully recovers the conditional distribution. Arithmetic coding can still be applied as the alphabet size is allowed to change while coding: what matters is that the conditional probabilities in each step sum to 1. Accordingly, we renormalize the top-k log-probabilities.

The Transformer models we trained specifically on enwik do not use any tokenization, except in \cref{sec:experiments:tokenization}. The reasoning above also holds, except that our models returned the full distribution over tokens, and not only the top-k.

\section{Additional Results}
\label{sec:additional-results}

\begin{figure}
    \begin{center}
        \begin{subfigure}{0.32\textwidth}
            \includegraphics[width=\textwidth]{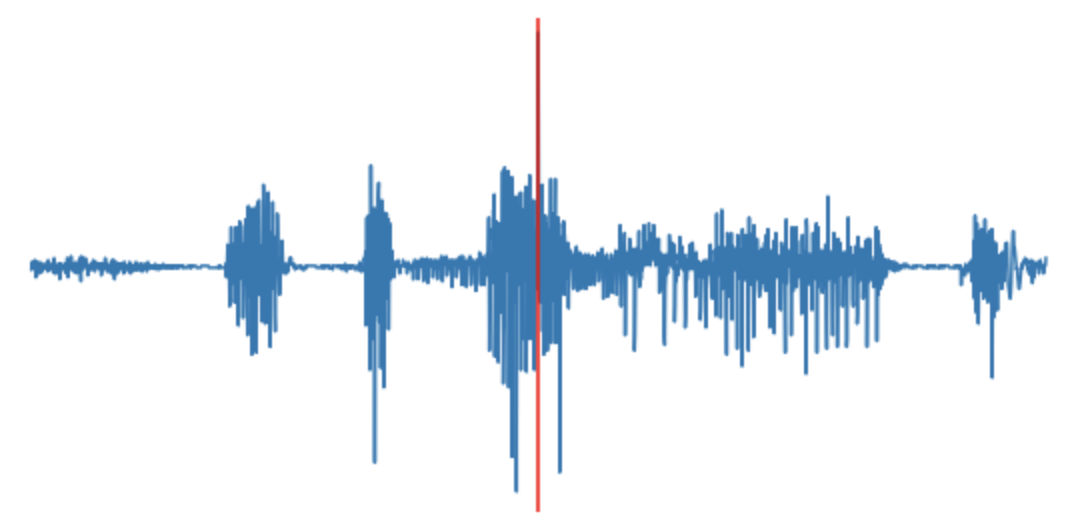}
            \caption{Original spectrogram}
        \end{subfigure}
        \hfill
        \begin{subfigure}{0.32\textwidth}
            \includegraphics[width=\textwidth]{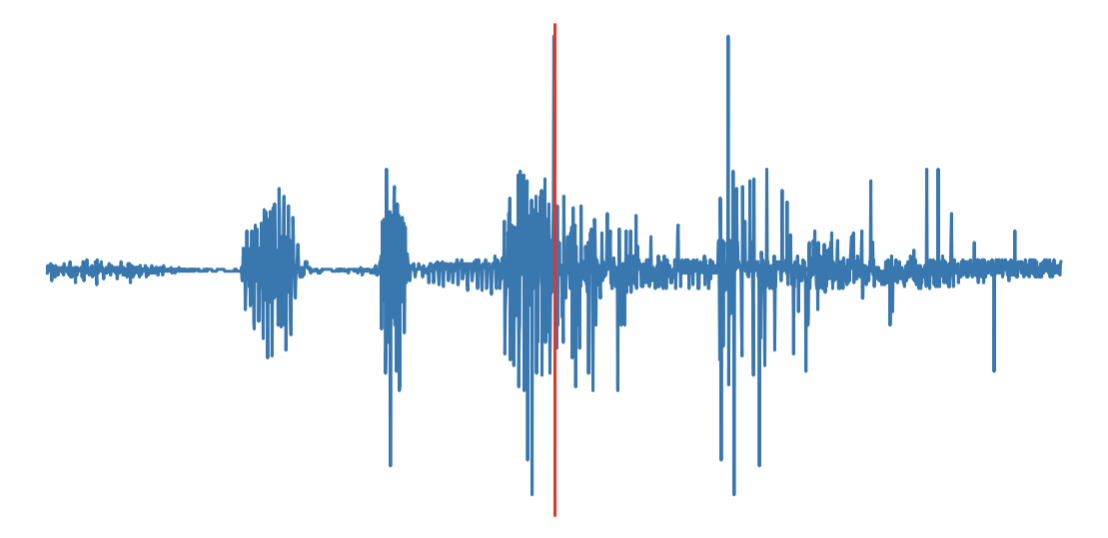}
            \caption{gzip}
        \end{subfigure}
        \hfill
        \begin{subfigure}{0.32\textwidth}
            \includegraphics[width=\textwidth]{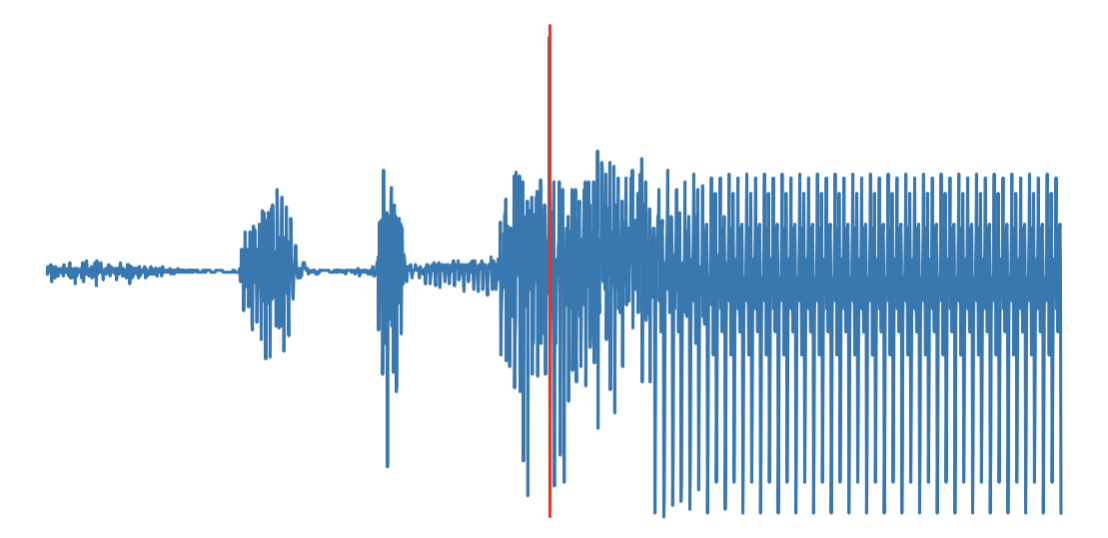}
            \caption{Chinchilla}
        \end{subfigure}
    \end{center}
    \caption{
        Compression-based generation for audio data.
        We condition gzip and Chinchilla on the first 1024 bytes of the base sequence (from LibriSpeech) and then sample the remaining 1024 bytes autoregressively.
        Chinchilla predictions exhibit a typical ``loop'' pattern of autoregressive generation.
    }
    \label{fig:generation-audio}
\end{figure}

\cref{fig:generation-text}, \cref{fig:generation-images} and \cref{fig:generation-audio} show data autoregressively generated by compressors, one step at a time. Note that for Chinchilla, we generate tokens (which size in bytes can vary) until we reach the length in bytes we desire. Also, note that gzip samples are biased, as explained in \cref{sec:experiments:generation}: looking 1 step ahead is not sufficient to get good samples, and it's likely that looking multiple steps ahead would improve the results. However, that's not the purpose of this paper, and we kept the simplest setup for all our compressors.

\begin{figure}
    \input{figures/text_generation}
    \caption{
        Compression-based generation for text data.
        We condition gzip and Chinchilla on a context text of size 1948 bytes (from enwik9) and then sample 100 bytes ($N$ tokens) autoregressively.
        Since Chinchilla employs a tokenizer, the sampled sequences will contain $N$ tokens, which do not necessarily decode to 100 bytes.  
        Chinchilla's predictions are significantly more coherent than gzip's.
    }
    \label{fig:generation-text}
\end{figure}

\begin{figure}
    \begin{center}
        \begin{subfigure}{0.32\textwidth}
            \includegraphics[width=\textwidth]{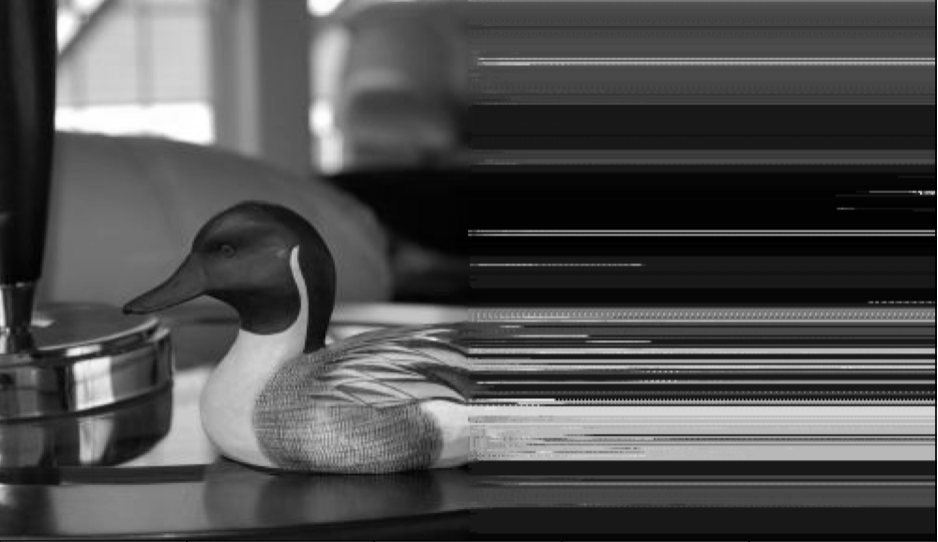}
            \caption{Chinchilla 1b}
        \end{subfigure}
        \hfill
        \begin{subfigure}{0.32\textwidth}
            \includegraphics[width=\textwidth]{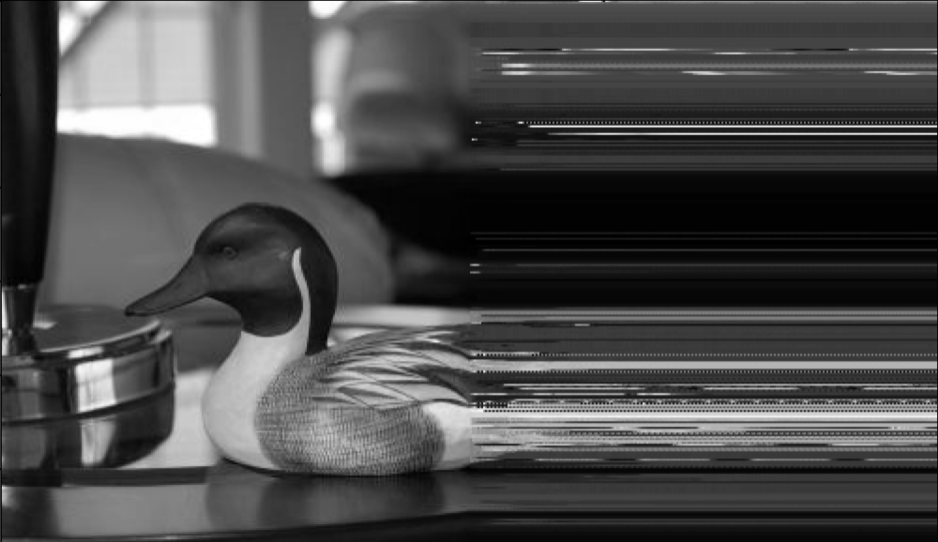}
            \caption{Chinchilla 7b}
        \end{subfigure}
        \hfill
        \begin{subfigure}{0.32\textwidth}
            \includegraphics[width=\textwidth]{figures/chinchilla70b_image_autoreg.png}
            \caption{Chinchilla 70b}
        \end{subfigure}
    \end{center}
    \caption{
        Compression-based generation for image data, for 3 Chinchilla models with different number of parameters.
        We condition the models on the first half of every row of the image (250 bytes) and then sample the remaining half (250 bytes) autoregressively.
    }
    \label{fig:generation-images-extra}
\end{figure}

\end{document}